\documentclass[10pt,twocolumn,letterpaper]{article}

\usepackage[pagenumbers]{cvpr} 

\usepackage{tabularx}
\usepackage{color}
\usepackage{algorithm}
\usepackage{algpseudocode}
\usepackage{pifont}
\usepackage{bm}
\usepackage{silence}
\usepackage{multirow}
\usepackage{amssymb}
\usepackage{amsmath}
\usepackage{marvosym}

\newcommand{\cmark}{\ding{52}}%
\newcommand{\xmark}{$\times$}%

\newcolumntype{Y}{>{\centering\arraybackslash}X}
\newcolumntype{Z}{>{\centering\arraybackslash}p{1.8cm}}
\definecolor{cvprblue}{rgb}{0.21,0.49,0.74}
\usepackage[pagebackref,breaklinks,colorlinks,allcolors=cvprblue]{hyperref}

\def\paperID{*****} 
\def\confName{CVPR}
\def\confYear{2026}

\title{CoPS: Conditional Prompt Synthesis for Zero-Shot Anomaly Detection}

\author{
Qiyu Chen$^{1,2}$ \quad
Zhen Qu$^{1,2}$ \quad
Wei Luo$^{4}$ \quad
Haiming Yao$^{4}$ \quad
Yunkang Cao$^{5}$ \quad
Yuxin Jiang$^{6}$\\
Yinan Duan$^{4}$ \quad
Huiyuan Luo$^{1}$ \quad
Chengkan Lv$^{1,3}$\textsuperscript{(\Letter)} \quad
Zhengtao Zhang$^{1,2,3}$\\
$^{1}$Institute of Automation, Chinese Academy of Sciences\\
$^{2}$School of Artificial Intelligence, University of Chinese Academy of Sciences\\
$^{3}$CASIVISION \quad
$^{4}$THU \quad
$^{5}$HNU \quad
$^{6}$HUST\\
{\tt\small \{chenqiyu2021,quzhen2022,huiyuan.luo,chengkan.lv,zhengtao.zhang\}@ia.ac.cn}\\
{\tt\small \{luow23,yhm22,dyn24\}@mails.tsinghua.edu.cn} \quad
{\tt\small caoyunkang@ieee.org} \quad
{\tt\small yuxinjiang@hust.edu.cn}
}

\begin{document}

\maketitle
\begin{abstract}    
    Recently, large pre-trained vision-language models have shown remarkable performance in zero-shot anomaly detection (ZSAD).
    With fine-tuning on a single auxiliary dataset,
    the model enables cross-category anomaly detection on diverse datasets covering industrial defects and medical lesions.
    Compared to manually designed prompts, prompt learning eliminates the need for expert knowledge and trial-and-error.
    However, it still faces the following challenges:
    (i) static learnable tokens struggle to capture the continuous and diverse patterns of normal and anomalous states,
    limiting generalization to unseen categories;
    (ii) fixed textual labels provide overly sparse category information, making the model prone to overfitting to a specific semantic subspace.
    To address these issues, we propose \underline{Co}nditional \underline{P}rompt \underline{S}ynthesis (CoPS),
    a novel framework that synthesizes dynamic prompts conditioned on visual features to enhance ZSAD performance.
    Specifically, we extract representative normal and anomaly prototypes from fine-grained patch features
    and explicitly inject them into prompts, enabling adaptive state modeling.
    Given the sparsity of class labels,
    we leverage a variational autoencoder to model semantic image features and implicitly fuse varied class tokens into prompts.
    Additionally, integrated with our spatially-aware alignment mechanism,
    extensive experiments demonstrate that
    CoPS surpasses state-of-the-art methods by 1.4\% in classification AUROC and 1.9\% in segmentation AUROC across 13 industrial and medical datasets.
    The code is available at \url{https://github.com/cqylunlun/CoPS}.
\end{abstract}
\begin{figure}[t]
    \centering
    \includegraphics[width=\linewidth]{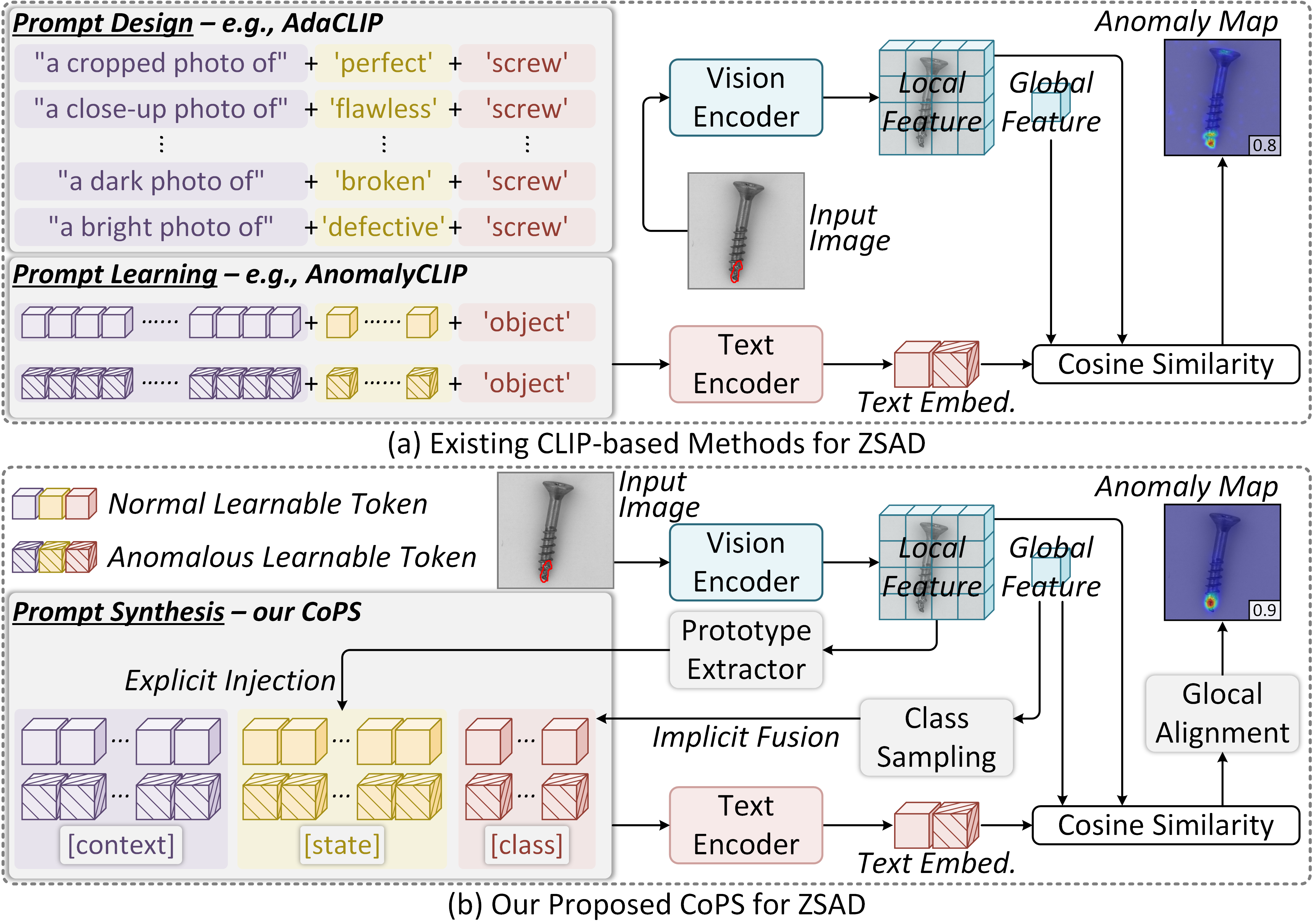}
    \caption{
        Comparison of existing CLIP-based methods and our proposed CoPS.
        (a) Existing methods rely on human-designed prompts or static learnable tokens.
        (b) Our method proposes explicit state token synthesis and implicit class token sampling to synthesize visual-conditioned prompts.
    }
    \label{fig:concept}
\end{figure}

\section{Introduction}
\label{sec:intro}

Anomaly detection (AD) aims to identify outliers that deviate from the normal data distribution, including
industrial defect detection \cite{gu2024anomalygpt,yao2024prior,chen2024unified} and
medical lesion localization \cite{huang2024adapting,lu2024anomaly,xiang2024exploiting}.
Classical AD methods \cite{zavrtanik2021draem,roth2022towards,liu2023simplenet}
are typically based on unsupervised learning using only normal samples and are trained to localize anomalies within known categories.
However, these methods are restricted to the categories seen during training and generalize poorly to novel classes with distribution shifts during testing.
This limitation poses significant challenges in real-world scenarios,
where collecting sufficient data across all categories is often impractical,
thereby motivating the need for zero-shot learning approaches.

To tackle this challenge, zero-shot anomaly detection (ZSAD) has been proposed to directly detect anomalies in previously unseen categories
by leveraging single auxiliary training data from disjoint classes.
Recently, several methods \cite{jeong2023winclip,chen2023april,chen2024clip,cao2024adaclip,zhu2025fine,xu2026mrad}
have demonstrated promising ZSAD capabilities by leveraging vision-language models (VLMs) pre-trained on large-scale image-text pairs.
Building upon the success of CLIP \cite{radford2021learning}, these methods design tailored prompts to adapt the model to downstream anomaly detection tasks.
As illustrated in Fig.~\ref{fig:concept}, each prompt can be decomposed into three components:
\underline{context} words, which describe the high-level imaging scene (e.g., ``a photo of'');
\underline{state} words, which specify the normal or anomalous condition (e.g., ``perfect''/``broken'');
and \underline{class} words, which capture the specific object category (e.g., ``screw'').

Existing methods can be broadly categorized into two types.
As depicted in Fig.~\ref{fig:concept}(a),
prompt design methods such as AdaCLIP~\cite{cao2024adaclip} rely on manually crafted template sets built from expert knowledge.
While these prompts are intuitive and interpretable, their construction requires extensive manual effort through repeated trial-and-error.
In contrast, prompt learning methods such as AnomalyCLIP~\cite{zhou2024anomalyclip} treat prompts as tunable parameters,
replacing the context and state words with static learnable tokens and using a single class-agnostic label for all samples.
However, this introduces two key limitations:
\emph{(i)} static learnable tokens fail to capture the continuous and diverse patterns of normal and anomalous states,
limiting generalization to unseen categories;
\emph{(ii)} fixed textual labels offer overly sparse semantic information,
making the model prone to overfitting within a narrow representation space.

To address these issues, we propose Conditional Prompt Synthesis (CoPS), a CLIP-based framework for ZSAD.
We assume that context words can be shared across images using static learnable tokens, since high-level imaging context is largely invariant across categories.
Meanwhile, the remaining prompt components are dynamically synthesized based on visual features, as shown in Fig.~\ref{fig:concept}(b).
Specifically, CoPS enhances prompt learning through two types of visual-conditioned token synthesis:
\emph{(i)} state words are injected with representative normal and anomaly prototypes,
which are extracted under a center constraint from fine-grained local features,
enabling better generalization through explicit state modeling;
\emph{(ii)} class words incorporate semantic global features sampled from a variational autoencoder (VAE) \cite{kingma2014auto},
enabling implicit label augmentation to improve prompt diversity.
Leveraging the observation that the distance between query feature and its nearest prototype also approximates anomaly state,
we introduce a distance-aware spatial attention mechanism to refine pixel-level text-image alignment.
Additionally, we employ a global-local (glocal) similarity interaction that further strengthens image-level alignment.

Our contributions are summarized as follows:
\begin{itemize}[leftmargin=1.1em]
    \item We propose CoPS, a novel framework built on CLIP to address the discretization of
    static learnable tokens and the sparsity of textual category labels,
    thereby improving ZSAD accuracy and generalization.
    \item We introduce two key modules to synthesize visual-conditioned prompts:
    an Explicit State Token Synthesis (ESTS) module for injecting normal and anomaly prototypes,
    and an Implicit Class Token Sampling (ICTS) module for fusing diverse semantic features.
    \item Integrated with a Spatially-Aware Glocal Alignment (SAGA) module,
    CoPS achieves state-of-the-art (SOTA) performance across 13 industrial and medical datasets under the ZSAD setting,
    improving classification and segmentation AUROC by 1.4\% and 1.9\%, respectively.
\end{itemize}
\section{Related Work}
\label{sec:related}

\subsection{Classical Anomaly Detection}
\label{sec:_cad}

In the classical AD setting,
the model is trained solely on normal samples and detects anomalies within the same known category during inference.
Specifically, reconstruction-based methods \cite{zavrtanik2021reconstruction,deng2022anomaly,lu2024anomaly,yao2024prior}
are widely used in classical AD, assuming the model accurately reconstructs normal samples but fails to reconstruct anomalies.
Anomalies are then detected by analyzing the residuals between the input and its reconstruction.
Embedding-based methods \cite{salehi2021multiresolution,roth2022towards,lee2022cfa,lei2023pyramidflow}
leverage pre-trained networks to extract features and compress normal features into a compact space.
Anomalies are then detected by measuring the distance between test features and normal clusters.
Since the aforementioned models cannot capture the anomalous distribution,
synthesis-based methods \cite{liu2023simplenet,chen2024unified,chen2024progressive,chen2025center,jiang2026anomagic,xu2025deltadeno}
augment training with synthetic anomalies to improve discrimination.
However, these methods rely on specific datasets,
limiting their applicability in real-world scenarios with distribution shifts between training and testing.

\begin{figure*}[t]
    \centering
    \includegraphics[width=\linewidth]{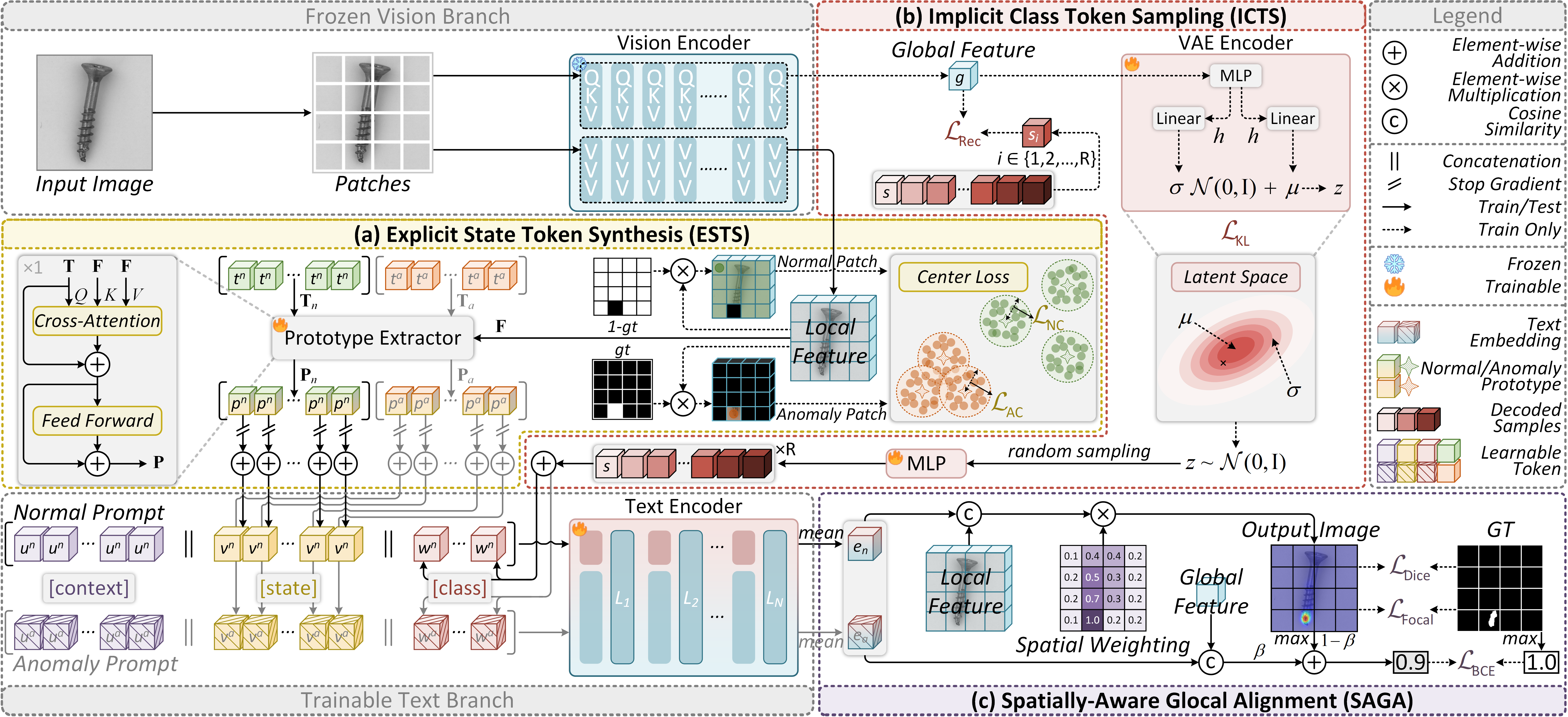}
    \caption{
        Schematic of CoPS.
        (a) ESTS extracts normal and anomaly prototypes from fine-grained local features and injects them into the state words.
        (b) ICTS leverages a VAE to model semantic global features and fuses decoded samples into the class words. 
        (c) SAGA applies distance-aware spatial attention to align textual embeddings with global and local visual features. 
    }
    \label{fig:schematic}
\end{figure*}

\subsection{Zero-Shot Anomaly Detection}
\label{sec:_zsad}

With the emergence of large pre-trained foundation models,
ZSAD on unseen categories has become feasible due to CLIP's strong cross-modal generalization and image-text alignment capability \cite{radford2021learning}.
Given the pivotal role of textual prompts in adapting CLIP to downstream tasks, ZSAD methods can be broadly categorized into two types.

\noindent
\textbf{Prompt design methods}
were introduced to model the relationship between normal and abnormal images using two sets of manually crafted textual templates.
WinCLIP \cite{jeong2023winclip} is the first to apply CLIP to ZSAD.
APRIL-GAN \cite{chen2023april} and CLIP-AD \cite{chen2024clip} introduce trainable adapters for fine-grained alignment.
AdaCLIP \cite{cao2024adaclip} further fine-tunes both text and vision encoders to enhance CLIP's ability to detect anomalies.

\noindent
\textbf{Prompt learning methods} have recently gained attention,
replacing fixed templates with learnable tokens to adaptively capture normal and anomalous semantics.
AnomalyCLIP \cite{zhou2024anomalyclip} builds on CoOp \cite{zhou2022learning} by making the context words learnable.
AdaptCLIP \cite{gao2025adaptclip} adopts an alternating training strategy between textual and visual representations for more stable improvement.
VCP-CLIP \cite{qu2024vcp} extends CoCoOp \cite{zhou2022conditional} by embedding image features into class words, eliminating the need for category labels.
Anomaly-OV \cite{xu2025towards} directly uses image features as class words and incorporates multimodal large language models (MLLMs) for anomaly reasoning.
FAPrompt \cite{zhu2025fine} synthesizes compound abnormality prompts by leveraging the mapped top-K abnormality-prior patch features.
However, these prompts are weakly constrained and semantically limited,
which hinders the generalization to unseen categories.
\section{Method}
\label{sec:method}

\subsection{Problem Definition}
\label{sec:_problem}

We follow the standard ZSAD setting, where the model is trained on seen categories $\mathcal{C}_s$ from an auxiliary training dataset and
evaluated on disjoint unseen categories $\mathcal{C}_u$ (i.e., $\mathcal{C}_s \cap \mathcal{C}_u = \varnothing$) without any adaptation.
Specifically, $\mathcal{C}_s$ are from an industrial dataset \(\mathcal{X}_\text{train}\),
while $\mathcal{C}_u$ are from industrial or medical datasets \(\mathcal{X}_\text{test}\) with distinct visual patterns.
The model outputs an image-level anomaly score and a pixel-level anomaly map for each image.

\subsection{Overview}
\label{sec:_overview}

Fig.~\ref{fig:schematic} illustrates the overall framework of CoPS, which consists of three key designs:
\emph{(i)} ESTS for state token synthesis, \emph{(ii)} ICTS for class token sampling, and \emph{(iii)} SAGA for glocal feature alignment.
We employ Transformer-based vision and text encoders from the pre-trained CLIP \cite{radford2021learning} as the backbone.
Given an input image \mbox{$\mathbf{X} \in \mathbb{R}^{h\times w\times 3}$},
the frozen vision encoder ${{\mathcal V}_\rho }$ first extracts semantic global feature \mbox{$\mathbf{g} \in \mathbb{R}^{C}$} and
fine-grained local feature \mbox{$\mathbf{F} \in \mathbb{R}^{HW \times C}$},
where \mbox{$H=h/p$} and \mbox{$W=w/p$} are the height and width for a patch size $p$, and $C$ is the embedding dimension.
Subsequently, ESTS introduces a prototype extractor ${{\mathcal P}_\theta }$ to generate \(M\) normal and anomaly prototypes
\mbox{$\mathbf{P}_n, \mathbf{P}_a\in\mathbb{R}^{M\times C}$}
from local feature $\mathbf{F}$, which are assembled into the dynamic state tokens.
ICTS then leverages a VAE ${{\mathcal E}_\psi }$ to parameterise the latent distribution of the global feature $\mathbf{g}$ and
draws $R$ decoded samples \mbox{$\mathbf{S}\in\mathbb{R}^{R\times C}$}, resulting in a dense set of class tokens.
Normal and anomaly prompt sets \mbox{$\{\tilde{\mathbf G}_i^n,\tilde{\mathbf G}_i^a\}_{i=1}^{R}$}
are constructed by concatenating the static context tokens with the dynamic state tokens and sampled class tokens.
To further enhance the image-text alignment, a learnable text encoder ${{\mathcal T}_\omega }$ is employed to
map these prompts into the textual embeddings \mbox{$\mathbf{e}_n, \mathbf{e}_a\in\mathbb{R}^{C}$}.
Finally, SAGA applies spatial attention to align \mbox{$\mathbf{e}_n, \mathbf{e}_a$} with the global feature $\mathbf{g}$ and local feature $\mathbf{F}$,
yielding an image-level anomaly score ${s}_{\text{cls}}$ and a pixel-level anomaly map $\bm{\mathcal{S}}_{\text{seg}}$.

\subsection{Learnable Dual-Prompt Construction}
\label{sec:_prompt}

Following \cite{qu2025bayesian}, we split text prompt into three components:
\underline{context} words, \underline{state} words, and \underline{class} words,
as shown in the lower-left corner of Fig.~\ref{fig:schematic}.
Recent work \cite{zhou2024anomalyclip} builds upon the prompt learning strategy \cite{zhou2022learning},
by replacing fixed context templates with learnable tokens.
However, they still manually craft a pair of state words (e.g., \textit{``good''/``damaged''}) and fix the class words to
class-agnostic text (e.g., \textit{``object''}), preventing either from being optimized and thereby limiting further gains in model generalization.
Accordingly, we introduce learnable tokens to jointly optimize all three components, producing the initial
normal and anomaly prompts \mbox{${{\bf{G}}^n},{{\bf{G}}^a} \in {\mathbb{R}^{L \times C}}$} with sequence length $L$:
\begin{align}
\mathbf{G}_{n} &=
  \left[ \mathbf{u}_{1}^{n} \right] \cdots \left[ \mathbf{u}_{K}^{n} \right]
  \parallel
  \left[ \mathbf{v}_{1}^{n} \right] \cdots \left[ \mathbf{v}_{M}^{n} \right]
  \parallel
  \left[ \mathbf{w}_{1}^{n} \right] \cdots \left[ \mathbf{w}_{N}^{n} \right]\notag\\
\mathbf{G}_{a} &=
  \left[ \mathbf{u}_{1}^{a} \right] \cdots \left[ \mathbf{u}_{K}^{a} \right]
  \parallel
  \left[ \mathbf{v}_{1}^{a} \right] \cdots \left[ \mathbf{v}_{M}^{a} \right]\,
  \parallel
  \left[ \mathbf{w}_{1}^{a} \right] \cdots \left[ \mathbf{w}_{N}^{a} \right]
    \label{eq:prompt_base}
\end{align}
where \mbox{${\bf{u}}_i^{n/a},{\bf{v}}_i^{n/a},{\bf{w}}_i^{n/a} \in {\mathbb{R}}^{C}$} are the learnable tokens in context, state, and class words, 
while $K$, $M$, and $N$ are their corresponding lengths.
The notation $n/a$ serves as a placeholder for either the normal or anomalous variable.

\subsection{Explicit State Token Synthesis (ESTS)}
\label{sec:_ests}

The normal and abnormal patterns of input images are continuous and diverse,
making it difficult for fixed state words to generalize in zero-shot settings.
To alleviate the overly discrete nature of the binary state text,
the ESTS module maps the most representative normal and abnormal local patches into two prototypes and explicitly injects them into the prompt,
enabling adaptive state modeling.

Since the original self-attention in CLIP builds relations among inconsistent semantic regions,
we adopt consistent self-attention (i.e., V-VV) in the frozen vision encoder ${{\mathcal V}_\rho }$ to
extract fine-grained local feature $\mathbf{F}$ without additional adaptation \cite{li2025closer}.
Fig.~\ref{fig:schematic}(a) introduces a prototype extractor $\mathcal{P}_\theta$ to
compute the cross-attention between \mbox{$\mathbf{F} \in \mathbb{R}^{HW \times C}$} and
normal and anomalous learnable tokens
\mbox{${{\bf{T}}_{n/a}} = \{ {\bf{t}}_1^{n/a},{\bf{t}}_2^{n/a}, \ldots ,{\bf{t}}_M^{n/a}|{\bf{t}}_m^{n/a} \in {\mathbb{R}}^{C}\}$},
yielding corresponding prototypes \mbox{$\mathbf{P}_n, \mathbf{P}_a\in\mathbb{R}^{M\times C}$} for each state:
\begin{align}
    {{\bf{Q}}_{n/a}} &= {{\bf{T}}_{n/a}}{{\bf{W}}^q},{{\bf{K}}_{n/a}} = {\bf{F}}{{\bf{W}}^k},{{\bf{V}}_{n/a}} = {\bf{F}}{{\bf{W}}^v}\notag\\
    {{\bf{T}}'_{n/a}} &= {\mathrm{Attention}}({{\bf{Q}}_{n/a}},{{\bf{K}}_{n/a}},{{\bf{V}}_{n/a}}) + {{\bf{T}}_{n/a}}\notag\\
    {{\bf{P}}_{n/a}} &= {\mathrm{FFN}}({{\bf{T}}'_{n/a}}) + {{\bf{T}}'_{n/a}}
    \label{eq:prototype}
\end{align}
where \mbox{${\bf{W}}^q,{\bf{W}}^k,{\bf{W}}^v \in {\mathbb{R}}^{C \times C}$} are the learnable projection matrices
for queries ${\bf{Q}}_{n}, {\bf{Q}}_{a} \in \mathbb{R}^{M \times C}$, keys ${\bf{K}}_{n}, {\bf{K}}_{a} \in \mathbb{R}^{HW \times C}$,
and values ${\bf{V}}_{n}, {\bf{V}}_{a} \in \mathbb{R}^{HW \times C}$.

\begin{figure}[t]
    \centering
    \includegraphics[width=\linewidth]{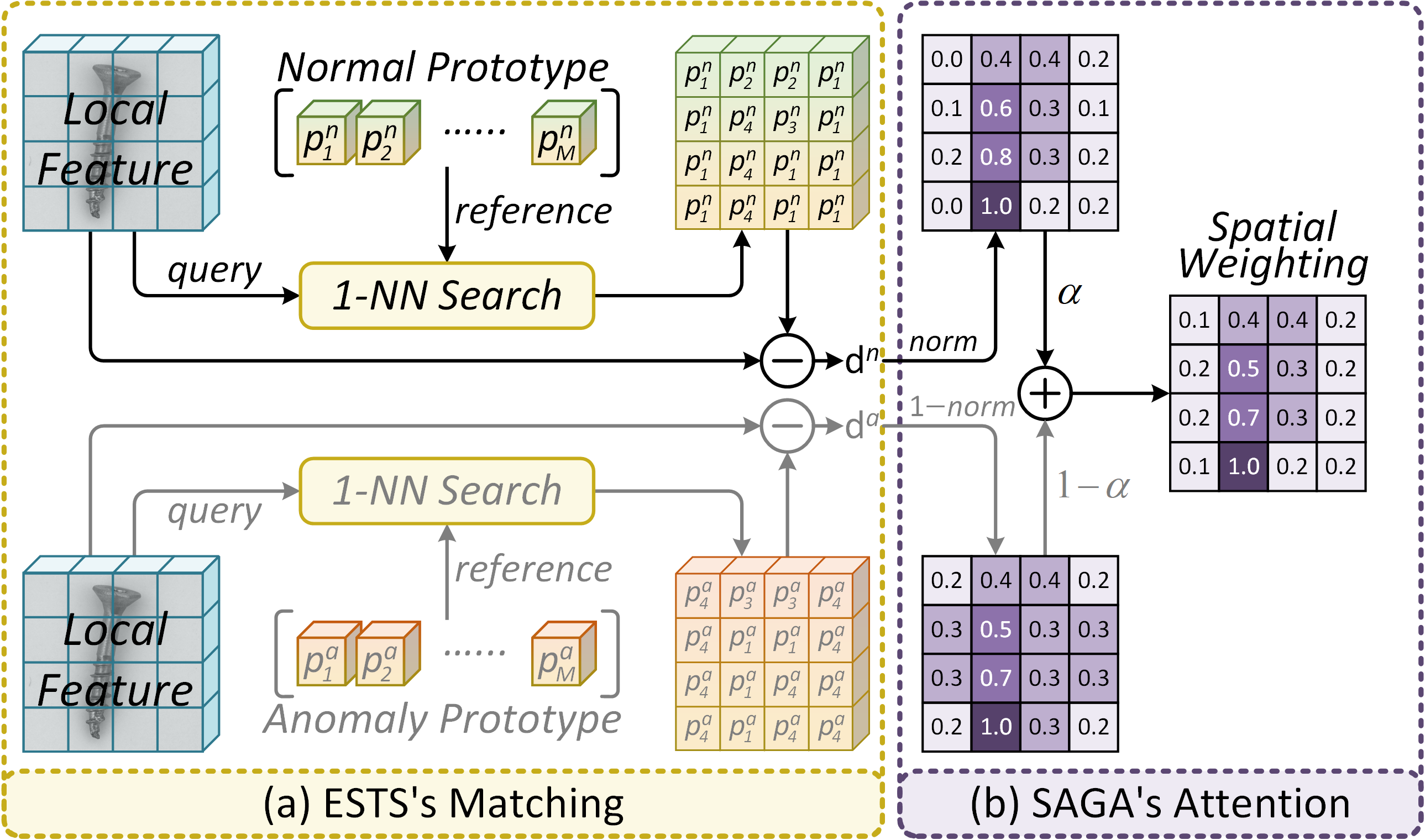}
    \caption{Pipeline of (a) ESTS's prototype matching and (b) SAGA's spatial attention.}
    \label{fig:pipeline}
\end{figure}

\noindent
\textbf{Center loss.}
To ensure that these two prototypes correspond to the negative and positive states \cite{luo2025inp},
we introduce normal center loss ${{\mathcal L}_{{\text{NC}}}}$ and abnormal center loss ${{\mathcal L}_{{\text{AC}}}}$ to
maximize the cosine similarity between each token \mbox{$\mathbf{f}_i \in \mathbb{R}^{C}$} of local feature $\mathbf{F}$ and
its nearest prototype \mbox{$\mathbf{p}_m \in \mathbf{P}$}:
\begin{align}
    d_i^{n/a} &=
    \min_{m \in \{ 1, \ldots ,M\} }
    \bigl( 1 - \cos(\mathbf{f}_i,\mathbf{p}^{n/a}_m) \bigr) \notag\\
    \mathcal{L}_{\text{ESTS}} &=
    \mathcal{L}_{\text{NC}} + \mathcal{L}_{\text{AC}}  \notag\\
    &= {\textstyle{1 \over {HW}}}
        \textstyle\sum_{i=1}^{HW} ({d_i^n \cdot (1 - {y_i}) + d_i^a \cdot {y_i}})
    \label{eq:l_ests}
\end{align}
where \mbox{${\bf{y}}=\{ {y_i}\} _{i = 1}^{HW}$} denotes the downsampled ground truth,
with \mbox{$y_i = 1$} for anomaly patches and \mbox{$y_i = 0$} for normal ones.
Fig.~\ref{fig:pipeline}(a) illustrates the prototype matching process in ESTS,
where \mbox{${\bf{d}}^{n/a} = \{ {d^{n/a}_i}\} _{i = 1}^{HW}$} is the distance between each query patch token
and its 1-nearest neighbor (1-NN) prototype.
As a result, the normal and anomaly prototypes \mbox{${\mathbf{P}}_{n/a} = \{ \mathbf{p}_m^{n/a} \}_{m = 1}^M$}
are respectively injected into the learnable state words of dual prompts in Eq.~(\ref{eq:prompt_base}).

\subsection{Implicit Class Token Sampling (ICTS)}
\label{sec:_icts}

Existing methods represent class words for different categories using
either multiple class-specific text labels \cite{cao2024adaclip} or a single class-agnostic text \cite{zhou2024anomalyclip}.
However, the learned semantic anchors in these models are overly sparse, making them prone to overfitting to a specific semantic subspace.
Therefore, the ICTS module leverages a VAE ${{\mathcal E}_\psi }$ to implicitly fuse rich semantic information
into the class words through distribution-consistent sampling.

The semantic global feature $\mathbf{g}$ is first extracted using the
original self-attention mechanism (i.e., Q-KV) in the frozen vision encoder ${{\mathcal V}_\rho }$.
As shown in Fig.~\ref{fig:schematic}(b), the VAE adopts a symmetric encoder--decoder architecture.
The global feature \mbox{$\mathbf{g} \in \mathbb{R}^{C}$} is fed into the VAE encoder ${q_{\psi '}}$ to
produce the latent variable \mbox{$\mathbf{z} \in \mathbb{R}^{C}$} via reparameterization
from the estimated mean \mbox{$\boldsymbol{\mu} \in \mathbb{R}^{C}$} and variance \mbox{$\boldsymbol{\sigma} \in \mathbb{R}^{C}$}:
\begin{align}
    {\bf{h}} &= \text{MLP}_e({\bf{g}}),{{\boldsymbol \mu }} = {\bf{h}}{{\bf{W}}^\mu } + {{\bf{b}}^\mu },\log {{{\boldsymbol \sigma }}^2} = {\bf{h}}{{\bf{W}}^\sigma } + {{\bf{b}}^\sigma }\notag\\
    \mathbf{z} &= \boldsymbol{\mu} + \boldsymbol{\sigma} \odot \boldsymbol{\epsilon}, \quad \boldsymbol{\epsilon} \sim \mathcal{N}(\mathbf{0}, \mathbf{I})
    \label{eq:encoder}
\end{align}
where \mbox{${\bf{W}}^\mu,{\bf{W}}^\sigma \in {\mathbb{R}}^{C \times C}$} and
\mbox{${\bf{b}}^\mu,{\bf{b}}^\sigma \in {\mathbb{R}}^{C}$} are the learnable weights of the linear layers,
and $\odot$ denotes the element-wise multiplication.
The latent variable $\mathbf{z}$ is then fed into the decoder ${p_{\psi''}}$ to obtain
the reconstructed global feature $\mathbf{s} \in \mathbb{R}^C$ (i.e., $\mathbf{s} = \text{MLP}_d(\mathbf{z})$).

\noindent
\textbf{VAE loss.}
To synthesize diverse and robust class tokens,
both the VAE encoder $q_{\psi '}$ and VAE decoder $p_{\psi ''}$ are optimized by maximizing the evidence lower bound (ELBO):
\begin{align}
    {{\mathcal L}_{{\text{ICTS}}}} &= {{\mathcal L}_{{\text{Rec}}}} + {{\mathcal L}_{{\text{KL}}}}\notag\\
    &= \mathbb{E}_{q_{\psi '}(\mathbf{z} | \mathbf{g})} [ -\log {p_{\psi ''}}(\mathbf{g} | \mathbf{z}) ] 
    + D_{\mathrm{KL}}(q_{\psi '}(\mathbf{z} | \mathbf{g}) \,\|\, p(\mathbf{z}))\notag\\
    &= \| {\mathbf{s}} - \mathbf{g} \|^2 + {\textstyle{1 \over {{2}}}}\textstyle\sum_{j=1}^{C}
    {( {\boldsymbol \mu _j^2 + \boldsymbol \sigma _j^2 \!-\! \log \boldsymbol \sigma _j^2 \!-\! 1})}
    \label{eq:l_icts}
\end{align}
where \mbox{$\mathcal{L}_{\text{Rec}}$} denotes the reconstruction loss and
\mbox{$\mathcal{L}_{\text{KL}}$} denotes the Kullback-Leibler (KL) divergence loss.
The prior $p(\mathbf{z})$ is assumed to follow a multivariate standard normal distribution $\mathcal{N}(\mathbf{0}, \mathbf{I})$.
We randomly sample $R$ reconstructed global features $\mathbf{s}_i$ from the latent space via the VAE decoder
to obtain a dense token set \mbox{${\mathbf{S}} = \{ \mathbf{s}_i \}_{i = 1}^R$}.
These are then fused into the learnable class words to generate
the complete dual prompt sets \mbox{$\{\tilde{\mathbf G}_i^n,\tilde{\mathbf G}_i^a\}_{i=1}^{R}$},
where each pair of dual prompts \mbox{$\tilde{\mathbf G}_i^n,\tilde{\mathbf G}_i^a \in \mathbb{R}^{L \times C}$} is constructed as:
\begin{align}
    {{\bf{\tilde G}}_i^n = \left[ {{\bf{u}}_1^n} \right] \cdots \left[ {{\bf{u}}_K^n} \right]
    \parallel \left[ {{\bf{v}}_1^n + {\bf{p}}_1^n} \right] \cdots \left[ {{\bf{v}}_M^n + {\bf{p}}_M^n} \right]}\notag\\
    {\parallel\; \left[ {{\bf{w}}_1^n + {{\bf{s}}_i}} \right]\; \cdots \;\left[ {{\bf{w}}_N^n + {{\bf{s}}_i}} \right]}\notag\\
    {{\bf{\tilde G}}_i^a = \left[ {{\bf{u}}_1^a} \right] \cdots \left[ {{\bf{u}}_K^a} \right]
    \parallel \,\left[ {{\bf{v}}_1^a + {\bf{p}}_1^a} \right] \cdots \left[ {{\bf{v}}_M^a + {\bf{p}}_M^a} \right]}\notag\\
    {\parallel\; \left[ {{\bf{w}}_1^a + {{\bf{s}}_i}} \right]\:\, \cdots \;\left[ {{\bf{w}}_N^a + {{\bf{s}}_i}} \right]}
    \label{eq:prompt}
\end{align}

\subsection{Spatially-Aware Glocal Alignment (SAGA)}
\label{sec:_saga}

Previous methods \cite{chen2023april,zhou2024anomalyclip} typically align textual embeddings independently
with global and local features to obtain classification and segmentation results.
However, this suffers from challenges in fine-grained and semantic text-image alignment.
To enhance generalization and accuracy, the SAGA module proposes prototype-guided spatial attention and glocal similarity interaction.

Fig.~\ref{fig:schematic}(c) shows that the dual prompt sets \mbox{$\{\tilde{\mathbf G}_i^n,\tilde{\mathbf G}_i^a\}_{i=1}^{R}$}
are first fed into the text encoder \mbox{${\mathcal T}_\omega$}, where the original prefix is replaced by learnable tokens.
The normal and abnormal textual embeddings \mbox{$\mathbf{e}_n, \mathbf{e}_a \in \mathbb{R}^{C}$}
are then obtained by averaging over sampling dimension.
To estimate the original likelihood of fine-grained patches and semantic image belonging to normal or anomalous categories,
we compute the cosine similarity between textual embeddings $\mathbf{e}_{n},\mathbf{e}_{a}$ and
local feature $\mathbf{F}$ as well as global feature $\mathbf{g}$:
\begin{align}
    {\bm{\mathcal S}_l^{n/a}} &= \left\{ {\frac{{\exp (\cos ({{\bf{e}}_{n/a}},{{\bf{f}}_i})/\tau )}}{{\exp (\cos ({{\bf{e}}_n},{{\bf{f}}_i})/\tau ) + \exp (\cos ({{\bf{e}}_a},{{\bf{f}}_i})/\tau )}}} \right\}_{i = 1}^{HW}\notag\\
    {{s}_g^{n/a}} &= \frac{{\exp (\cos ({{\bf{e}}_{n/a}},{\bf{g}})/\tau )}}{{\exp (\cos ({{\bf{e}}_n},{\bf{g}})/\tau ) + \exp (\cos ({{\bf{e}}_a},{\bf{g}})/\tau )}}
    \label{eq:initial_score}
\end{align}
where \mbox{$\tau$} is a temperature hyperparameter and \mbox{${\bf{f}}_i \in \mathbb{R}^{C}$} is the $i$-th patch token of $\mathbf{F}$,
while ${\bm{\mathcal S}_l^{n/a}} \in \mathbb{R}^{HW}$ denotes the initial local similarity map and
$s_g^{n/a} \in \mathbb{R}$ denotes the initial global similarity score.

Since a local token is more likely to be normal when its distance $\mathbf{d}^n$ to normal prototypes is small
and its distance $\mathbf{d}^a$ to anomaly prototypes is large (and vice versa),
we introduce prototype-based spatial attention to further enhance text-image alignment.
As illustrated in Fig.~\ref{fig:pipeline}(b), $\mathbf{d}^n$ and $\mathbf{d}^a$ are combined to compute the spatial weighting mask:
\begin{equation}
    {\bf{M}_\text{mask}} = \alpha  \cdot \frac{{{{\bf{d}}^n}}}{{\left\| {{{\bf{d}}^n}} \right\|}} + (1 - \alpha )\left( {1 - \frac{{{{\bf{d}}^a}}}{{\left\| {{{\bf{d}}^a}} \right\|}}} \right)
    \label{eq:spatial}
\end{equation}
where distance coefficient $\alpha$ balances the influence of normal distance $\mathbf{d}^n$ and anomaly distance $\mathbf{d}^a$.
The spatial weighting mask \mbox{$\mathbf{M}_\text{mask} \in \mathbb{R}^{HW}$} is applied to refine the local similarity map,
which is then aggregated to compute the refined global similarity score:
\begin{align}
    {\bm{\hat \mathcal S}}^{n/a}_l &= {\bm{\mathcal S}}^{n/a}_l \odot {\bf{M}}_\text{mask}\notag\\
    \hat s_g^{n/a} &= \beta \cdot s_g^{n/a} + (1 - \beta ) \cdot \max ({\bm{\hat \mathcal S}}_l^{n/a})
    \label{eq:score}
\end{align}
where glocal coefficient \mbox{$\beta$} controls the trade-off between local and global scores.

\noindent
\textbf{Glocal loss.}
Following \cite{zhou2024anomalyclip},
we enhance the model's binary discriminative ability at both pixel and image levels by introducing a glocal loss defined as:
\begin{align}
    {\mathcal L}_{\text{SAGA}} &= {{\mathcal L}_{{\text{Dice}}}} + {{\mathcal L}_{{\text{Focal}}}} + {{\mathcal L}_{{\text{BCE}}}}\notag\\
    &= \text{Dice}({\bm{\tilde {\mathcal S}}}_l^a, {\bf{Y}}) + \text{Dice}({\bm{\tilde {\mathcal S}}}_l^n, {\bf{1}} - {\bf{Y}}) \notag\\
    &\, + \text{Focal}([{\bm{\tilde {\mathcal S}}}_l^n, {\bm{\tilde {\mathcal S}}}_l^a], {\bf{Y}}) 
    + \text{BCE}({\hat s}_g^a, \max ({\bf{Y}}))
    \label{eq:l_saga}
\end{align}
where \mbox{${\bf{Y}} \in \mathbb{R}^{h \times w}$} denotes the full-size ground truth,
and \mbox{$\bm{\tilde{\mathcal S}}_l^{n/a} \in \mathbb{R}^{h \times w}$} is the reshaped and interpolated version of
the refined local similarity map \mbox{$\bm{\hat{\mathcal S}}_l^{n/a} \in \mathbb{R}^{HW}$}.

\subsection{Training and Inference}
\label{sec:_training}

Since the prototype extractor \( {\mathcal P}_\theta \) already enforces explicit prototype alignment through the center loss,
the ESTS module is not optimized using SAGA's glocal loss.
In contrast, the ICTS module implicitly samples from the global feature distribution and can be trained jointly with the VAE loss and glocal loss.
Finally, the overall training objective is:
\begin{equation}
    {\mathcal J} = \mathop {\min }\limits_\theta  {{\mathcal L}_{{\mathrm{ESTS}}}} + \mathop {\min }\limits_\psi  {{\mathcal L}_{{\mathrm{ICTS}}}} + \mathop {\min }\limits_{\psi ,\omega,\varphi} {{\mathcal L}_{{\mathrm{SAGA}}}}
    \label{eq:l_all}
\end{equation}
where $\theta$, $\psi$, $\omega$, and $\varphi$ are the learnable parameters of
the ESTS module, the ICTS module, the text encoder, and the initial dual prompts, respectively.

During inference, the vision encoder first extracts the global feature $\mathbf{g}$ and local feature $\mathbf{F}$ from input image $\mathbf{X}$.
The state prototypes $\mathbf{P}$ are obtained from $\mathbf{F}$ via ESTS, and the class token sets $\mathbf{S}$ are sampled from the standard Gaussian prior via ICTS.
These are then fused into the initial dual prompts to synthesize complete normal and anomaly prompt sets.
Subsequently, the text encoder generates textual embeddings $\mathbf{e}$, which are aligned with $\mathbf{g}$ and $\mathbf{F}$ via SAGA.
Finally, the image-level anomaly score ${s}_{\text{cls}}$ is given by the refined global similarity score $\hat s_g^a$,
and the pixel-level anomaly map $\bm{\mathcal{S}}_{\text{seg}}$ is obtained by applying Gaussian filtering to the refined local similarity map $\bm{\hat{\mathcal S}}_l^a$.

\begin{table*}[t]\small
  \centering
  \caption{Performance comparison of various SOTA methods on industrial and medical datasets under ZSAD setting, as measured by I-AUROC\% / I-AP\%.
   The best results are highlighted in bold, and the second-best results are underlined.}
    \begin{tabularx}{\textwidth}{p{1.4cm}|p{1.6cm}|YZYY|ZY|Y}
    \noalign{\hrule height 0.4mm}
    \multicolumn{1}{l|}{\multirow{3}{*}{Domain~$\downarrow$}} & Method~$\rightarrow$ & \multicolumn{4}{c|}{Prompt Design} & \multicolumn{3}{c}{Prompt Learning} \\
\cline{2-9}    \multicolumn{1}{l|}{} & \multirow{2}{*}{Dataset~$\downarrow$} & WinCLIP & APRIL-GAN & CLIP-AD & AdaCLIP & AnomalyCLIP & FAPrompt & \multirow{2}{*}{\textbf{CoPS}} \\
    \multicolumn{1}{l|}{} &       & CVPR'23 & CVPRw'23 & IJCAI'24 & ECCV'24 & ICLR'24 & ICCV'25 &  \\
    \hline
    \multicolumn{1}{l|}{\multirow{5}{*}{Industrial}} & MVTec-AD & 91.8 / 95.1 & 86.1 / 93.5 & 89.8 / 95.3 & \underline{92.0} / \underline{96.4} & 91.5 / 96.2 & 91.9 / 95.9 & \textbf{95.0} / \textbf{97.8} \\
    \multicolumn{1}{l|}{} & VisA  & 78.1 / 77.5 & 78.0 / 81.4 & 79.8 / 84.3 & 83.0 / 84.9 & 82.1 / 85.4 & \underline{84.0} / \underline{86.7} & \textbf{85.4} / \textbf{88.0} \\
    \multicolumn{1}{l|}{} & BTAD  & 83.3 / 84.1 & 74.2 / 71.7 & 85.8 / 85.2 & 91.6 / \underline{92.4} & 89.1 / 91.1 & \underline{92.3} / 90.1 & \textbf{93.6} / \textbf{94.9} \\
    \multicolumn{1}{l|}{} & MPDD  & 61.5 / 69.2 & \underline{76.8} / \textbf{82.9} & 71.6 / 76.3 & 76.4 / 80.4 & 73.7 / 76.5 & 76.5 / 80.3 & \textbf{78.6} / \underline{81.1} \\
    \multicolumn{1}{l|}{} & DTD-Synth & \underline{95.0} / \underline{97.9} & 83.9 / 93.6 & 91.5 / 96.8 & 92.8 / 97.0 & 94.5 / 97.7 & 94.9 / 97.6 & \textbf{95.2} / \textbf{98.1} \\
    \hline
    \multicolumn{1}{l|}{\multirow{3}{*}{Medical}} & HeadCT & 83.7 / 81.6 & 89.3 / 89.6 & 93.8 / 92.2 & 93.4 / 92.2 & \underline{95.3} / 95.2 & \textbf{96.1} / \underline{96.6} & \textbf{96.1} / \textbf{97.1} \\
    \multicolumn{1}{l|}{} & BrainMRI & 92.0 / 90.7 & 89.6 / 84.5 & 92.8 / 85.5 & 94.9 / 94.2 & \underline{96.1} / 92.3 & 95.9 / \underline{96.2} & \textbf{97.4} / \textbf{97.6} \\
    \multicolumn{1}{l|}{} & Br35H & 80.5 / 82.2 & 93.1 / 92.9 & 96.0 / 95.5 & 95.7 / 95.7 & \underline{97.3} / \underline{96.1} & 97.2 / \underline{96.1} & \textbf{98.7} / \textbf{98.5} \\
    \hline
    \multicolumn{2}{l|}{Average Performance} & 83.2 / 84.8 & 83.9 / 86.3 & 87.6 / 88.9 & 90.0 / 91.7 & 90.0 / 91.3 & \underline{91.1} / \underline{92.4} & \textbf{92.5} / \textbf{94.1} \\
    \noalign{\hrule height 0.4mm}
    \end{tabularx}
  \label{tab:image_level_compare}%
\end{table*}%

\section{Experiments}
\label{sec:exper} 

\subsection{Experimental Setup}
\label{sec:_setup}

\noindent
\textbf{Datasets.}
We evaluate the ZSAD performance of our proposed method on 13 publicly available datasets from industrial and medical domains.
Specifically, we employ five widely used industrial datasets:
MVTec-AD \cite{bergmann2019mvtec}, VisA \cite{zou2022spot}, BTAD \cite{mishra2021vt}, MPDD \cite{jezek2021deep}, and DTD-Synthetic \cite{aota2023zero}.
Additionally, we utilize eight medical datasets:
HeadCT \cite{salehi2021multiresolution}, BrainMRI \cite{kanade2015brain}, Br35H \cite{hamada2020brain},
ISIC \cite{codella2018skin}, CVC-ColonDB \cite{tajbakhsh2015automated}, CVC-ClinicDB \cite{bernal2015wm},
Kvasir \cite{jha2019kvasir}, and Endo \cite{hicks2021endotect}.
Since the categories in VisA are disjoint from those in the other datasets,
we use VisA as the auxiliary training set and evaluate on all remaining datasets.
For VisA evaluation, the model is fine-tuned on the MVTec-AD dataset.
All experiments are conducted under the same evaluation protocol to ensure a fair comparison.
More detailed descriptions about the datasets can be found in Appendix Sec.~A.

\noindent
\textbf{Evaluation metrics.}
To evaluate the ZSAD performance at both image and pixel levels,
we employ the area under the receiver operating characteristic (AUROC) and average precision (AP) as evaluation metrics.
AUROC and AP are denoted as I-AUROC/I-AP for the image-level classification and P-AUROC/P-AP for the pixel-level segmentation.

\noindent
\textbf{Implementation details.}
Following previous works \cite{zhou2024anomalyclip,cao2024adaclip,zhu2025fine},
we adopt the publicly available CLIP (ViT-L/14@336px) pre-trained by OpenAI~\cite{radford2021learning}. 
Input images are resized to \mbox{$518\times518$}, and the final layer of the vision encoder is used to extract global and local embeddings.
For ESTS, the context token length $K$, state token length $M$, and class token length $N$ are set to 6, 6, and 2, respectively.
For ICTS, the sampling count $R$ is set to 10.
For SAGA, the distance coefficient $\alpha$ and glocal coefficient $\beta$ are set to 0.3 and 0.9, respectively.
CoPS is trained using the Adam optimizer for 10 epochs with an initial learning rate of 0.001 and a batch size of 8.
All experiments are conducted on a system equipped with a single NVIDIA GeForce RTX 3090 GPU and an Intel Xeon Gold 6226R CPU.
Additional implementation details can be found in Appendix Sec.~B.

\begin{table*}[t]\small
  \centering
  \caption{Performance comparison of various SOTA methods on industrial and medical datasets under ZSAD setting, as measured by P-AUROC\% / P-AP\%.
   The best results are highlighted in bold, and the second-best results are underlined.}
    \begin{tabularx}{\textwidth}{p{1.4cm}|p{1.6cm}|YZYY|ZY|Y}
    \noalign{\hrule height 0.4mm}
    \multicolumn{1}{l|}{\multirow{3}{*}{Domain~$\downarrow$}} & Method~$\rightarrow$ & \multicolumn{4}{c|}{Prompt Design} & \multicolumn{3}{c}{Prompt Learning} \\
\cline{2-9}    \multicolumn{1}{l|}{} & \multirow{2}{*}{Dataset~$\downarrow$} & WinCLIP & APRIL-GAN & CLIP-AD & AdaCLIP & AnomalyCLIP & FAPrompt & \multirow{2}{*}{\textbf{CoPS}} \\
    \multicolumn{1}{l|}{} &       & CVPR'23 & CVPRw'23 & IJCAI'24 & ECCV'24 & ICLR'24 & ICCV'25 &  \\
    \hline
    \multicolumn{1}{l|}{\multirow{5}{*}{Industrial}} & MVTec-AD & 85.1 / 18.0 & 87.6 / \underline{40.8} & 89.8 / 40.0 & 86.8 / 38.1 & \underline{91.1} / 34.5 & 90.4 / 34.6 & \textbf{93.4} / \textbf{41.9} \\
    \multicolumn{1}{l|}{} & VisA  & 79.6 / 5.00 & 94.2 / 25.7 & 95.0 / \underline{26.3} & 95.1 / \textbf{29.2} & \underline{95.5} / 21.3 & \textbf{95.7} / 21.4 & \textbf{95.7} / 23.4 \\
    \multicolumn{1}{l|}{} & BTAD  & 71.4 / 11.2 & 91.3 / 32.9 & 93.1 / \textbf{46.7} & 87.7 / 36.6 & 93.3 / 42.0 & \underline{93.7} / 40.4 & \textbf{94.6} / \underline{42.6} \\
    \multicolumn{1}{l|}{} & MPDD  & 71.2 / 14.1 & 95.2 / 24.9 & \underline{96.7} / 26.3 & 95.2 / \underline{28.5} & 96.2 / 28.0 & 93.9 / 26.2 & \textbf{97.5} / \textbf{30.9} \\
    \multicolumn{1}{l|}{} & DTD-Synth & 82.5 / 11.6 & 96.6 / \textbf{67.3} & 97.1 / \underline{62.3} & 94.1 / 52.8 & \underline{97.6} / 52.4 & \underline{97.6} / 53.5 & \textbf{98.4} / 58.5 \\
    \hline
    \multicolumn{1}{l|}{\multirow{5}{*}{Medical}} & ISIC  & 83.5 / 62.4 & 85.8 / 69.8 & 81.6 / 65.5 & 85.4 / 70.6 & 88.4 / 74.4 & \underline{90.4} / \underline{77.7} & \textbf{93.8} / \textbf{85.6} \\
    \multicolumn{1}{l|}{} & ColonDB & 64.8 / 14.3 & 78.4 / 23.2 & 80.3 / 23.7 & 79.3 / 26.2 & 81.9 / 31.3 & \underline{85.2} / \underline{36.7} & \textbf{85.6} / \textbf{37.2} \\
    \multicolumn{1}{l|}{} & ClinicDB & 70.7 / 19.4 & \underline{86.0} / 38.8 & 85.8 / 39.0 & 84.3 / 36.0 & 85.9 / 42.2 & 85.5 / \underline{45.1} & \textbf{88.8} / \textbf{49.9} \\
    \multicolumn{1}{l|}{} & Kvasir & 69.8 / 27.5 & 80.2 / 42.4 & 82.5 / 46.2 & 79.4 / 43.8 & 81.8 / 42.5 & \underline{84.0} / \underline{46.6} & \textbf{85.8} / \textbf{51.5} \\
    \multicolumn{1}{l|}{} & Endo  & 68.2 / 23.8 & 84.1 / 47.9 & 85.6 / 51.7 & 84.0 / 44.8 & 86.3 / 50.4 & \underline{88.4} / \underline{55.8} & \textbf{90.0} / \textbf{58.7} \\
    \hline
    \multicolumn{2}{l|}{Average Performance} & 74.7 / 20.7 & 87.9 / 41.4 & 88.8 / 42.8 & 87.1 / 40.7 & 89.8 / 41.9 & \underline{90.5} / \underline{43.8} & \textbf{92.4} / \textbf{48.0} \\
    \noalign{\hrule height 0.4mm}
    \end{tabularx}%
  \label{tab:pixel_level_compare}%
\end{table*}%

\begin{figure*}[t]
    \centering
    \includegraphics[width=\linewidth]{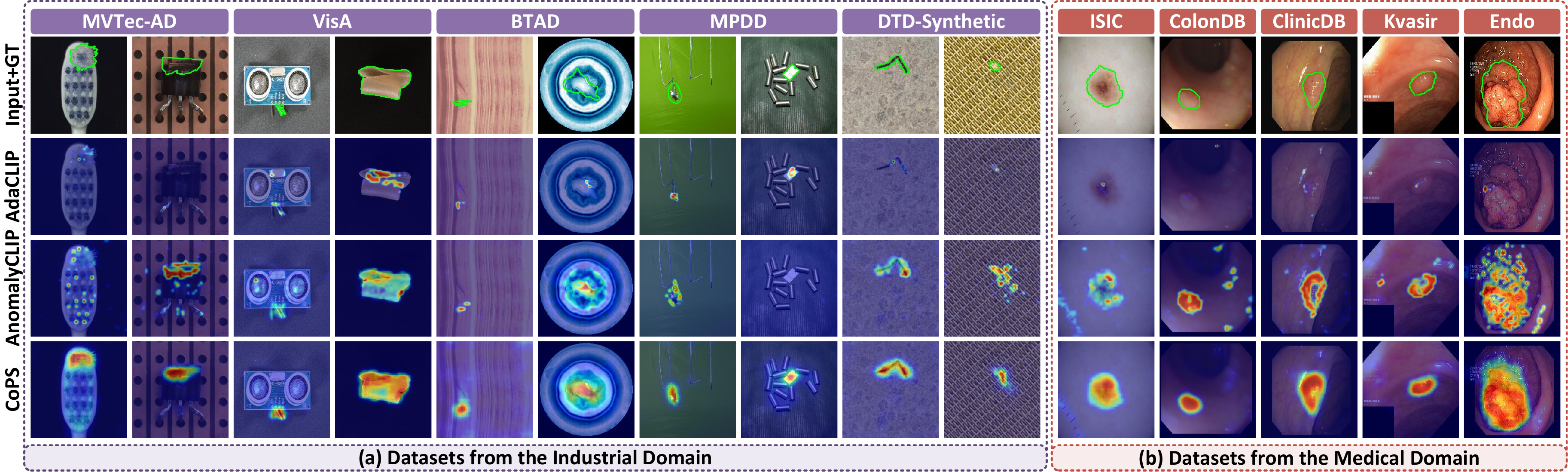}
    \caption{Qualitative comparison of CoPS against representative methods (AdaCLIP \cite{cao2024adaclip} and AnomalyCLIP \cite{zhou2024anomalyclip}) across multiple datasets.}
    \label{fig:visual}
\end{figure*}

\subsection{Comparison with State-of-the-Art}
\label{sec:_comparative}

The proposed CoPS is compared with six SOTA methods on 13 industrial and medical datasets.
Prompt design-based methods include:
WinCLIP \cite{jeong2023winclip}, APRIL-GAN \cite{chen2023april}, CLIP-AD \cite{chen2024clip}, and AdaCLIP \cite{cao2024adaclip},
while prompt learning-based methods include:
AnomalyCLIP \cite{zhou2024anomalyclip} and FAPrompt \cite{zhu2025fine}.
For fair comparison, all methods use the same backbone, input resolution, and evaluation protocol.

\begin{table*}[t]\small
  \centering
  \caption{Performance ablation of different components in CoPS on industrial and medical datasets under ZSAD setting, as measured by I-AUROC\% / P-AUROC\%.
  The best results are highlighted in bold, and the second-best results are underlined.}
    \begin{tabularx}{0.98\textwidth}{p{1.8cm}|Y|YYY|YYY|>{\centering\arraybackslash}p{1.4cm}}
    \noalign{\hrule height 0.4mm}
    \multirow{2}{*}{Module~$\downarrow$} & \multicolumn{7}{c|}{Model Variants}   &  \\
\cline{2-8}          & (A)   & (B)   & (C)   & (D)   &  (E)  &  (F)  &  (G) & \multirow{-2}{*}{\textbf{Ours}} \\
    \hline
    \textbf{ESTS} &  \xmark     & \cmark     &  \xmark     &  \xmark     & \cmark     & \cmark     & \xmark     & \cmark \\
    \textbf{ICTS} &  \xmark     & \xmark     &  \cmark     &  \xmark     &  \cmark    & \xmark     & \cmark     & \cmark \\
    \textbf{SAGA} &  \xmark     & \xmark     &  \xmark     &  \cmark     & \xmark     & \cmark     & \cmark     & \cmark \\
    \hline
    MVTec-AD & 91.1 / 92.0 & 93.2 / 93.0 & 91.7 / 92.0 & 92.4 / 92.2 & \underline{94.5} / \underline{93.3} & 93.5 / \underline{93.3} & 92.6 / 92.2 & \textbf{95.0 / 93.4} \\
    BTAD  & 91.6 / 92.4 & 93.0 / 93.7 & 92.0 / 92.9 & 91.9 / 94.0 & \underline{93.4} / 93.8 & 93.3 / \underline{94.4} & 92.4 / 94.2 & \textbf{93.6 / 94.6} \\
    \hline
    HeadCT & 94.2 / \,~~-~~~~ & 94.8 / \,~~-~~~~ & 94.3 / \,~~-~~~~ & 94.6 / \,~~-~~~~ & 95.0 / \,~~-~~~~ & 95.5 / \,~~-~~~~ & \underline{95.6} / \,~~-~~~~ & \textbf{96.1 / \,~~-~~~~} \\
    Endo  & ~~~-~~\, / 88.5 & ~~~-~~\, / 89.0 & ~~~-~~\, / 88.6 & ~~~-~~\, / 88.9 & ~~~-~~\, / 89.3 & ~~~-~~\, / \underline{89.7} & ~~~-~~\, / 89.1 & \textbf{~~~-~~\, / 90.0} \\
    \hline
    Average & 92.3 / 91.0 & 93.7 / 91.9 & 92.7 / 91.2 & 93.0 / 91.7 & \underline{94.3} / 92.1 & 94.1 / \underline{92.5} & 93.5 / 91.8 & \textbf{94.9 / 92.7} \\
    \noalign{\hrule height 0.4mm}
    \end{tabularx}%
  \label{tab:ablation}%
\end{table*}%

\noindent
\textbf{Quantitative results.}
As shown in Tab.~\ref{tab:image_level_compare},
CoPS achieves the best or second-best performance on classification tasks across all industrial and medical datasets,
with an average improvement of 1.4\% in I-AUROC and 1.7\% in I-AP.
Benefiting from their tunable prompt representations, prompt learning-based methods generally outperform prompt design-based approaches on downstream tasks.
CoPS further improves generalization by injecting constrained and sampled visual conditions into the synthesized prompts, leading to superior performance.
Furthermore, Tab.~\ref{tab:pixel_level_compare} shows that CoPS also achieves state-of-the-art performance on segmentation tasks,
with average improvements of 1.9\% in P-AUROC and 4.2\% in P-AP.
Due to the extremely small defect regions in VisA and DTD-Synthetic, the AP metric is highly sensitive to over-detection.
Although CoPS does not achieve the highest P-AP on these two datasets, it still surpasses our baseline AnomalyCLIP by 2.1\% and 6.1\%, respectively.
Compared with AnomalyCLIP, CoPS benefits from spatial weighting, which further refines anomaly localization at the target regions.
More quantitative results are presented in Appendix Sec.~D.

\noindent
\textbf{Qualitative results.}
As shown in Fig.~\ref{fig:visual}, we present the visualization of our CoPS and
two representative ZSAD methods AdaCLIP \cite{cao2024adaclip} and AnomalyCLIP \cite{zhou2024anomalyclip} on 15 categories from 10 datasets with pixel-level annotations. 
Fig.~\ref{fig:visual}(a) shows that CoPS achieves precise localization on industrial datasets while reducing false positives on object foregrounds.
In contrast, AdaCLIP produces numerous missed detections due to the high variance of anomaly scores and limited ability to capture large defect regions.
Fig.~\ref{fig:visual}(b) demonstrates that CoPS provides more comprehensive coverage of lesion regions on medical datasets.
Meanwhile, AnomalyCLIP often yields over-detection because its normal and anomalous prompts lack continuity and diversity, leading to excessive activation in normal areas.
These results indicate that our method synthesizes more effective prompts through prototype extraction and class sampling,
which provide strong generalization for both industrial and medical ZSAD tasks.
More qualitative results are presented in Appendix Sec.~E.

\subsection{Ablation Study}
\label{sec:_ablation}

This section presents ablation studies to further assess the impact of each component and hyperparameters on CoPS.

\noindent
\textbf{Influence of different components.}
Tab.~\ref{tab:ablation} reports ablation results of seven model variants against the complete model
on four industrial and medical datasets to quantify the contribution of each component.
All model variants retain learnable prompt tokens.
Specifically, variant (A) serves as the baseline, similar to the AnomalyCLIP \cite{zhou2024anomalyclip} framework.
Variants (B-D) independently incorporate ESTS, ICTS, and SAGA into (A), each yielding performance gains.
Notably, ESTS brings the largest improvement at both image and pixel levels, followed by SAGA and ICTS.
Variants (E-G) are derived by removing SAGA, ICTS, and ESTS from the complete model, respectively.
Both ICTS and SAGA further enhance classification and segmentation performance,
while removing ESTS leads to the most significant drop by 1.4\% in I-AUROC and 0.9\% in P-AUROC.
These results highlight the effectiveness of prototype extraction, class sampling, and glocal alignment.

\begin{figure}[t]
    \centering
    \includegraphics[width=\linewidth]{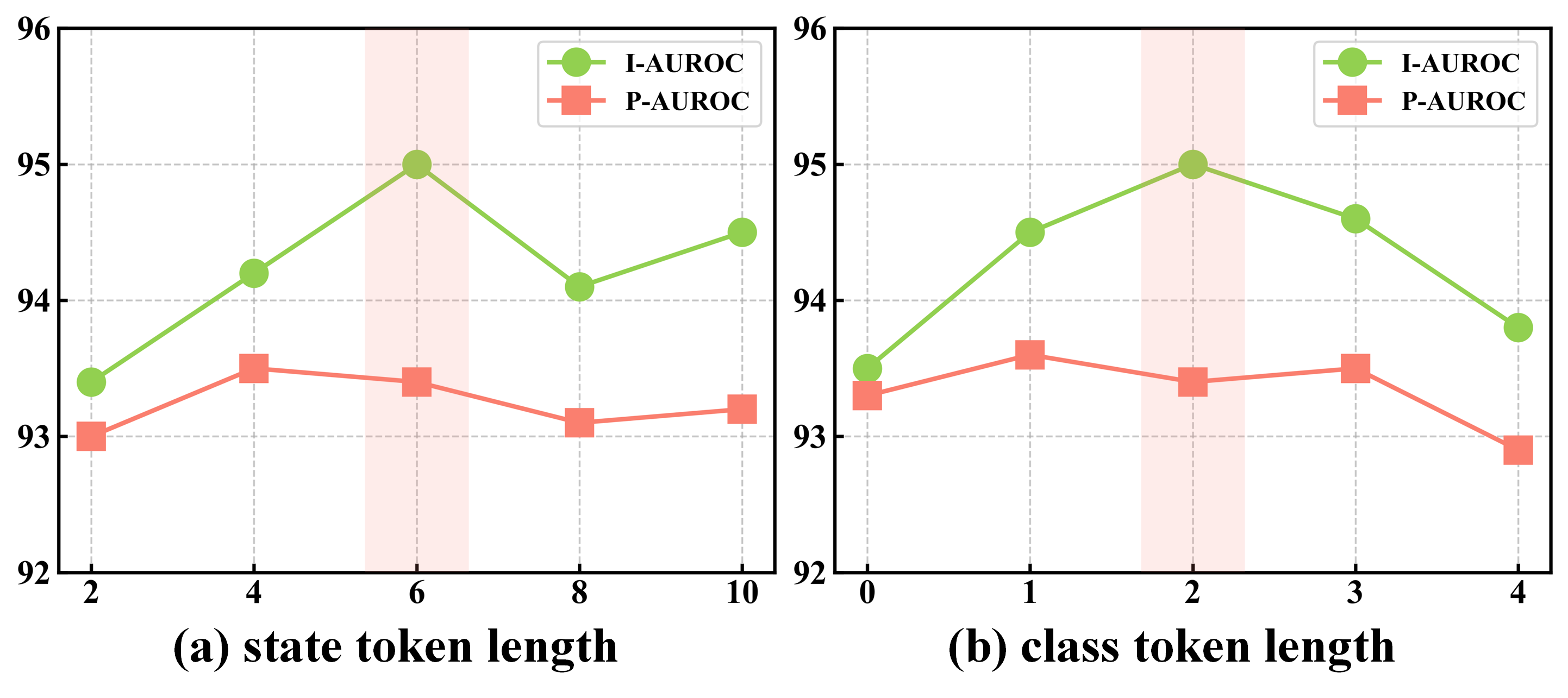}
    \caption{Performance ablation of state and class token length.}
    \label{fig:linechart}
\end{figure}

\noindent
\textbf{Influence of learnable token length.}
Fig.~\ref{fig:linechart}(a) illustrates the effect of varying the state token length $M$,
which corresponds to the number of normal and abnormal prototypes in ESTS.
Performance increases from \mbox{$M=2$} to \mbox{$M=6$}, suggesting that a larger prototype set better captures state diversity.
However, further increasing $M$ to 8 and 10 results in a slight performance drop, likely caused by overfitting.
Fig.~\ref{fig:linechart}(b) shows the effect of varying class token length $N$ in ICTS.
Performance increases from \mbox{$N=0$} to \mbox{$N=2$}, suggesting that small-scale class sampling enhances semantic diversity.
Similarly, performance drops when $N$ exceeds 2, indicating diminishing returns from additional class words.

\begin{table}[t]\small
  \centering
  \caption{Performance ablation of sampling count $R$.
  The selected sampling count is underlined.}
    \begin{tabularx}{0.45\textwidth}{p{0.3cm}|Y>{\centering\arraybackslash}p{0.75cm}Y>{\centering\arraybackslash}p{0.75cm}|>{\centering\arraybackslash}p{0.75cm}}
    \noalign{\hrule height 0.4mm}
    R     & I-AUROC & I-AP  & P-AUROC & P-AP  & 1/FPS \\
    \hline
    0     & 93.5\%  & 97.2\%  & 93.3\%  & 41.8\%  & 140ms \\
    1     & 94.5\%  & 97.6\%  & 93.5\%  & 41.9\%  & 144ms \\
    5     & 94.9\%  & 97.8\%  & 93.4\%  & 42.0\%  & 154ms \\
    \underline{10}    & \underline{95.0\%}  & \underline{97.8\%}  & \underline{93.4\%}  & \underline{41.9\%}  & \underline{168ms} \\
    15    & 94.9\%  & 97.7\%  & 93.4\%  & 41.8\%  & 179ms \\
    20    & 94.9\%  & 97.7\%  & 93.5\%  & 41.9\%  & 185ms \\
    \noalign{\hrule height 0.4mm}
    \end{tabularx}%
  \label{tab:sampling}%
\end{table}%

\noindent
\textbf{Influence of sampling count.}
Tab.~\ref{tab:sampling} presents the ablation of sampling count $R$ in ICTS.
Increasing $R$ from 0 to 10 leads to consistent gains at the image level and relatively minor gains at the pixel level,
indicating that class sampling enriches prompt diversity.
Since performance saturates while inference time increases beyond \mbox{$R=10$},
this value offers a balanced trade-off between accuracy and efficiency.
We select \mbox{$R=10$} as the default setting.

\begin{figure}[t]
    \centering
    \includegraphics[width=\linewidth]{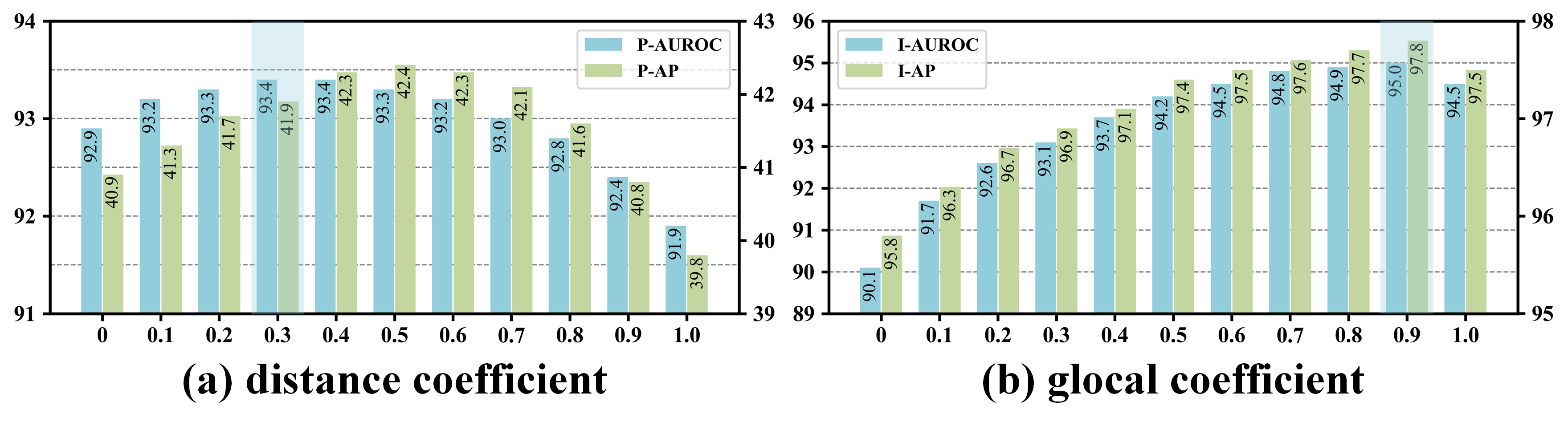}
    \caption{Performance ablation of distance and glocal coefficient.}
    \label{fig:barchart}
\end{figure}

\noindent
\textbf{Influence of weight coefficients.}
As shown in Fig.~\ref{fig:barchart}(a), increasing the distance coefficient $\alpha$ in SAGA first improves performance and then leads to a decline.
We select \mbox{$\alpha=0.3$} to balance the contributions of normal and abnormal prototype distances in the spatial weighting mask.
As shown in Fig.~\ref{fig:barchart}(b), increasing the glocal coefficient $\beta$ consistently improves both I-AUROC and I-AP, with peak performance at $\beta = 0.9$.
This trend highlights the effectiveness of glocal alignment in integrating global and local semantics for anomaly detection.
However, performance drops notably at $\beta = 1.0$, indicating that relying solely on global information without local score hinders generalization.
More ablation study can be found in Appendix Sec.~C.

\section{Conclusion}
\label{sec:conclu}

In this work, we propose CoPS, a conditional prompt synthesis framework for ZSAD.
By explicitly injecting state prototypes and implicitly fusing class semantics, CoPS synthesizes dynamic prompts conditioned on visual features,
addressing the limitations of discrete state tokens and sparse class labels.
Furthermore, our spatially-aware glocal alignment enhances generalization and accuracy.
These designs enable CoPS to achieve SOTA zero-shot anomaly classification and segmentation on 13 industrial and medical datasets.

\noindent
\textbf{Limitations \& Future Work.}
Although CoPS shows strong performance in detecting structural anomalies across diverse industrial and medical datasets,
it remains limited in handling anomalies that require high-level semantic reasoning.
This limitation stems from the lack of deep understanding of object relations and functional semantics within current framework.
To this end, future work will integrate MLLMs to enhance CoPS's logical anomaly detection capability and improve generalization to complex scenarios.

\section*{Acknowledgements}
\label{sec:acknow}
This work is supported by National Natural Science Foundation of China under Grants 62303461 and 62303458.

\small
\bibliographystyle{ieeenat_fullname}
\bibliography{main}

@article{chen2024progressive,
  title={Progressive boundary guided anomaly synthesis for industrial anomaly detection},
  author={Chen, Qiyu and Luo, Huiyuan and Gao, Han and Lv, Chengkan and Zhang, Zhengtao},
  journal={IEEE Transactions on Circuits and Systems for Video Technology},
  year={2025},
  volume={35},
  number={2},
  pages={1193-1208},
}

@inproceedings{chen2024unified,
  title={A unified anomaly synthesis strategy with gradient ascent for industrial anomaly detection and localization},
  author={Chen, Qiyu and Luo, Huiyuan and Lv, Chengkan and Zhang, Zhengtao},
  booktitle={European Conference on Computer Vision},
  pages={37--54},
  year={2024},
}

@article{chen2025center,
  title={Center-aware residual anomaly synthesis for multiclass industrial anomaly detection},
  author={Chen, Qiyu and Luo, Huiyuan and Yao, Haiming and Luo, Wei and Qu, Zhen and Lv, Chengkan and Zhang, Zhengtao},
  journal={IEEE Transactions on Industrial Informatics}, 
  year={2025},
  volume={21},
  number={9},
  pages={7276-7286},
}

@inproceedings{cao2024adaclip,
  title={Adaclip: Adapting clip with hybrid learnable prompts for zero-shot anomaly detection},
  author={Cao, Yunkang and Zhang, Jiangning and Frittoli, Luca and Cheng, Yuqi and Shen, Weiming and Boracchi, Giacomo},
  booktitle={European Conference on Computer Vision},
  pages={55--72},
  year={2024},
}

@inproceedings{qu2024vcp,
  title={Vcp-clip: A visual context prompting model for zero-shot anomaly segmentation},
  author={Qu, Zhen and Tao, Xian and Prasad, Mukesh and Shen, Fei and Zhang, Zhengtao and Gong, Xinyi and Ding, Guiguang},
  booktitle={European Conference on Computer Vision},
  pages={301--317},
  year={2024}
}

@inproceedings{zhou2024anomalyclip,
  title={Anomalyclip: Object-agnostic prompt learning for zero-shot anomaly detection},
  author={Zhou, Qihang and Pang, Guansong and Tian, Yu and He, Shibo and Chen, Jiming},
  booktitle={International Conference on Learning Representations},
  year={2024},
}

@article{xu2026mrad,
  title={MRAD: Zero-shot anomaly detection with memory-driven retrieval},
  author={Xu, Chaoran and Lv, Chengkan and Chen, Qiyu and Zhang, Feng and Zhang, Zhengtao},
  journal={International Conference on Learning Representations},
  year={2026}
}

@inproceedings{radford2021learning,
  title={Learning transferable visual models from natural language supervision},
  author={Radford, Alec and Kim, Jong Wook and Hallacy, Chris and Ramesh, Aditya and Goh, Gabriel and Agarwal, Sandhini and Sastry, Girish and Askell, Amanda and Mishkin, Pamela and Clark, Jack and others},
  booktitle={International Conference on Machine Learning},
  pages={8748--8763},
  year={2021},
}

@inproceedings{zhou2022conditional,
  title={Conditional prompt learning for vision-language models},
  author={Zhou, Kaiyang and Yang, Jingkang and Loy, Chen Change and Liu, Ziwei},
  booktitle={Proceedings of the IEEE/CVF Conference on Computer Vision and Pattern Recognition},
  pages={16816--16825},
  year={2022},
}

@inproceedings{qu2025bayesian,
  title={Bayesian prompt flow learning for zero-shot anomaly detection},
  author={Qu, Zhen and Tao, Xian and Gong, Xinyi and Qu, Shichen and Chen, Qiyu and Zhang, Zhengtao and Wang, Xingang and Ding, Guiguang},
  booktitle={Proceedings of the IEEE/CVF Conference on Computer Vision and Pattern Recognition},
  pages={30398--30408},
  year={2025},
}

@article{zhou2022learning,
  title={Learning to prompt for vision-language models},
  author={Zhou, Kaiyang and Yang, Jingkang and Loy, Chen Change and Liu, Ziwei},
  journal={International Journal of Computer Vision},
  volume={130},
  number={9},
  pages={2337--2348},
  year={2022},
}

@article{li2025closer,
  title={A closer look at the explainability of contrastive language-image pre-training},
  author={Li, Yi and Wang, Hualiang and Duan, Yiqun and Zhang, Jiheng and Li, Xiaomeng},
  journal={Pattern Recognition},
  volume={162},
  pages={111409},
  year={2025},
}

@inproceedings{zhu2025fine,
  title={Fine-grained abnormality prompt learning for zero-shot anomaly detection},
  author={Zhu, Jiawen and Ong, Yew-Soon and Shen, Chunhua and Pang, Guansong},
  booktitle={Proceedings of the IEEE/CVF International Conference on Computer Vision},
  pages={22241--22251},
  year={2025}
}

@article{luo2025inp,
  title={Inp-former++: Advancing universal anomaly detection via intrinsic normal prototypes and residual learning},
  author={Luo, Wei and Yao, Haiming and Cao, Yunkang and Chen, Qiyu and Gao, Ang and Shen, Weiming and Yu, Wenyong},
  journal={arXiv preprint arXiv:2506.03660},
  year={2025}
}

@inproceedings{bergmann2019mvtec,
  title={Mvtec ad-a comprehensive real-world dataset for unsupervised anomaly detection},
  author={Bergmann, Paul and Fauser, Michael and Sattlegger, David and Steger, Carsten},
  booktitle={Proceedings of the IEEE/CVF Conference on Computer Vision and Pattern Recognition},
  pages={9592-9600},
  year={2019}
}

@inproceedings{zou2022spot,
  title={Spot-the-difference self-supervised pre-training for anomaly detection and segmentation},
  author={Zou, Yang and Jeong, Jongheon and Pemula, Latha and Zhang, Dongqing and Dabeer, Onkar},
  booktitle={European Conference on Computer Vision},
  pages={392-408},
  year={2022}
}

@inproceedings{mishra2021vt,
  title={Vt-adl: A vision transformer network for image anomaly detection and localization},
  author={Mishra, Pankaj and Verk, Riccardo and Fornasier, Daniele and Piciarelli, Claudio and Foresti, Gian Luca},
  booktitle={International Symposium on Industrial Electronics},
  pages={01--06},
  year={2021}
}

@inproceedings{jezek2021deep,
  title={Deep learning-based defect detection of metal parts: Evaluating current methods in complex conditions},
  author={Jezek, Stepan and Jonak, Martin and Burget, Radim and Dvorak, Pavel and Skotak, Milos},
  booktitle={International Congress on Ultra Modern Telecommunications and Control Systems and Workshops},
  pages={66--71},
  year={2021}
}

@inproceedings{aota2023zero,
  title={Zero-shot versus many-shot: Unsupervised texture anomaly detection},
  author={Aota, Toshimichi and Tong, Lloyd Teh Tzer and Okatani, Takayuki},
  booktitle={Proceedings of the IEEE/CVF Winter Conference on Applications of Computer Vision},
  pages={5564--5572},
  year={2023}
}

@article{kanade2015brain,
  title={Brain tumor detection using mri images},
  author={Kanade, Pranita Balaji and Gumaste, PP},
  journal={Brain},
  volume={3},
  number={2},
  pages={146--150},
  year={2015}
}

@misc{hamada2020brain,
  title={Brain tumor detection},
  author={Ahmed Hamada},
  year={2020},
  howpublished={www.kaggle.com/datasets/ahmedhamada0/brain-tumor-detection},
  note={Online}
}

@inproceedings{codella2018skin,
  title={Skin lesion analysis toward melanoma detection: A challenge at the 2017 international symposium on biomedical imaging (isbi), hosted by the international skin imaging collaboration (isic)},
  author={Codella, Noel CF and Gutman, David and Celebi, M Emre and Helba, Brian and Marchetti, Michael A and Dusza, Stephen W and Kalloo, Aadi and Liopyris, Konstantinos and Mishra, Nabin and Kittler, Harald and others},
  booktitle={International Symposium on Biomedical Imaging},
  pages={168--172},
  year={2018}
}

@article{tajbakhsh2015automated,
  title={Automated polyp detection in colonoscopy videos using shape and context information},
  author={Tajbakhsh, Nima and Gurudu, Suryakanth R and Liang, Jianming},
  journal={IEEE Transactions on Medical Imaging},
  volume={35},
  number={2},
  pages={630--644},
  year={2015}
}

@article{bernal2015wm,
  title={Wm-dova maps for accurate polyp highlighting in colonoscopy: Validation vs. saliency maps from physicians},
  author={Bernal, Jorge and S{\'a}nchez, F Javier and Fern{\'a}ndez-Esparrach, Gloria and Gil, Debora and Rodr{\'\i}guez, Cristina and Vilari{\~n}o, Fernando},
  journal={Computerized Medical Imaging and Graphics},
  volume={43},
  pages={99--111},
  year={2015}
}

@inproceedings{jha2019kvasir,
  title={Kvasir-seg: A segmented polyp dataset},
  author={Jha, Debesh and Smedsrud, Pia H and Riegler, Michael A and Halvorsen, P{\aa}l and De Lange, Thomas and Johansen, Dag and Johansen, H{\aa}vard D},
  booktitle={International Conference on Multimedia Modeling},
  pages={451--462},
  year={2019}
}

@inproceedings{hicks2021endotect,
  title={The endoTect 2020 challenge: Evaluation and comparison of classification, segmentation and inference time for endoscopy},
  author={Hicks, Steven A and Jha, Debesh and Thambawita, Vajira and Halvorsen, P{\aa}l and Hammer, Hugo L and Riegler, Michael A},
  booktitle={International Conference on Pattern Recognition},
  pages={263--274},
  year={2021}
}

@inproceedings{chen2024clip,
  title={Clip-ad: A language-guided staged dual-path model for zero-shot anomaly detection},
  author={Chen, Xuhai and Zhang, Jiangning and Tian, Guanzhong and He, Haoyang and Zhang, Wuhao and Wang, Yabiao and Wang, Chengjie and Liu, Yong},
  booktitle={International Joint Conference on Artificial Intelligence},
  pages={17--33},
  year={2024}
}

@inproceedings{gao2025adaptclip,
  title={Adaptclip: Adapting clip for universal visual anomaly detection},
  author={Gao, Bin-Bin and Zhou, Yue and Yan, Jiangtao and Cai, Yuezhi and Zhang, Weixi and Wang, Meng and Liu, Jun and Liu, Yong and Wang, Lei and Wang, Chengjie},
  booktitle={Proceedings of the AAAI Conference on Artificial Intelligence},
  year={2026}
}

@inproceedings{jeong2023winclip,
  title={Winclip: Zero-/few-shot anomaly classification and segmentation},
  author={Jeong, Jongheon and Zou, Yang and Kim, Taewan and Zhang, Dongqing and Ravichandran, Avinash and Dabeer, Onkar},
  booktitle={Proceedings of the IEEE/CVF Conference on Computer Vision and Pattern Recognition},
  pages={19606--19616},
  year={2023}
}

@article{chen2023april,
  title={April-gan: A zero-/few-shot anomaly classification and segmentation method for cvpr 2023 vand workshop challenge tracks 1\&2: 1st place on zero-shot ad and 4th place on few-shot ad},
  author={Chen, Xuhai and Han, Yue and Zhang, Jiangning},
  journal={arXiv preprint arXiv:2305.17382},
  year={2023}
}

@article{zavrtanik2021reconstruction,
  title={Reconstruction by inpainting for visual anomaly detection},
  author={Zavrtanik, Vitjan and Kristan, Matej and Sko{\v{c}}aj, Danijel},
  journal={Pattern Recognition},
  volume={112},
  pages={107706},
  year={2021}
}

@inproceedings{deng2022anomaly,
  title={Anomaly detection via reverse distillation from one-class embedding},
  author={Deng, Hanqiu and Li, Xingyu},
  booktitle={Proceedings of the IEEE/CVF Conference on Computer Vision and Pattern Recognition},
  pages={9737--9746},
  year={2022}
}

@article{yao2024prior,
  title={Prior normality prompt transformer for multiclass industrial image anomaly detection},
  author={Yao, Haiming and Cao, Yunkang and Luo, Wei and Zhang, Weihang and Yu, Wenyong and Shen, Weiming},
  journal={IEEE Transactions on Industrial Informatics},
  volume={20},
  number={10},
  pages={11866--11876},
  year={2024}
}

@inproceedings{salehi2021multiresolution,
  title={Multiresolution knowledge distillation for anomaly detection},
  author={Salehi, Mohammadreza and Sadjadi, Niousha and Baselizadeh, Soroosh and Rohban, Mohammad H and Rabiee, Hamid R},
  booktitle={Proceedings of the IEEE/CVF Conference on Computer Vision and Pattern Recognition},
  pages={14902--14912},
  year={2021}
}

@inproceedings{roth2022towards,
  title={Towards total recall in industrial anomaly detection},
  author={Roth, Karsten and Pemula, Latha and Zepeda, Joaquin and Sch{\"o}lkopf, Bernhard and Brox, Thomas and Gehler, Peter},
  booktitle={Proceedings of the IEEE/CVF Conference on Computer Vision and Pattern Recognition},
  pages={14318--14328},
  year={2022}
}

@article{lee2022cfa,
  title={Cfa: Coupled-hypersphere-based feature adaptation for target-oriented anomaly localization},
  author={Lee, Sungwook and Lee, Seunghyun and Song, Byung Cheol},
  journal={IEEE Access},
  volume={10},
  pages={78446--78454},
  year={2022}
}

@inproceedings{lei2023pyramidflow,
  title={Pyramidflow: High-resolution defect contrastive localization using pyramid normalizing flow},
  author={Lei, Jiarui and Hu, Xiaobo and Wang, Yue and Liu, Dong},
  booktitle={Proceedings of the IEEE/CVF Conference on Computer Vision and Pattern Recognition},
  pages={14143--14152},
  year={2023}
}

@inproceedings{liu2023simplenet,
  title={Simplenet: A simple network for image anomaly detection and localization},
  author={Liu, Zhikang and Zhou, Yiming and Xu, Yuansheng and Wang, Zilei},
  booktitle={Proceedings of the IEEE/CVF Conference on Computer Vision and Pattern Recognition},
  pages={20402--20411},
  year={2023}
}

@inproceedings{zavrtanik2021draem,
  title={Draem-a discriminatively trained reconstruction embedding for surface anomaly detection},
  author={Zavrtanik, Vitjan and Kristan, Matej and Sko{\v{c}}aj, Danijel},
  booktitle={Proceedings of the IEEE/CVF International Conference on Computer Vision},
  pages={8330--8339},
  year={2021}
}

@inproceedings{xu2025towards,
  title={Towards zero-shot anomaly detection and reasoning with multimodal large language models},
  author={Xu, Jiacong and Lo, Shao-Yuan and Safaei, Bardia and Patel, Vishal M and Dwivedi, Isht},
  booktitle={Proceedings of the IEEE/CVF International Conference on Computer Vision},
  pages={20370--20382},
  year={2025}
}

@inproceedings{huang2024adapting,
  title={Adapting visual-language models for generalizable anomaly detection in medical images},
  author={Huang, Chaoqin and Jiang, Aofan and Feng, Jinghao and Zhang, Ya and Wang, Xinchao and Wang, Yanfeng},
  booktitle={Proceedings of the IEEE/CVF Conference on Computer Vision and Pattern Recognition},
  pages={11375--11385},
  year={2024}
}

@article{lu2024anomaly,
  title={Anomaly detection for medical images using heterogeneous auto-encoder},
  author={Lu, Shuai and Zhang, Weihang and Zhao, He and Liu, Hanruo and Wang, Ningli and Li, Huiqi},
  journal={IEEE Transactions on Image Processing},
  volume={33},
  pages={2770--2782},
  year={2024}
}

@article{xiang2024exploiting,
  title={Exploiting structural consistency of chest anatomy for unsupervised anomaly detection in radiography images},
  author={Xiang, Tiange and Zhang, Yixiao and Lu, Yongyi and Yuille, Alan and Zhang, Chaoyi and Cai, Weidong and Zhou, Zongwei},
  journal={IEEE Transactions on Pattern Analysis and Machine Intelligence},
  volume={46},
  number={9},
  pages={6070--6081},
  year={2024},
}

@inproceedings{gu2024anomalygpt,
  title={Anomalygpt: Detecting industrial anomalies using large vision-language models},
  author={Gu, Zhaopeng and Zhu, Bingke and Zhu, Guibo and Chen, Yingying and Tang, Ming and Wang, Jinqiao},
  booktitle={Proceedings of the AAAI Conference on Artificial Intelligence},
  volume={38},
  pages={1932--1940},
  year={2024}
}

@inproceedings{kingma2014auto,
  title={Auto-encoding variational bayes},
  author={Kingma, Diederik P and Welling, Max},
  booktitle={International Conference on Learning Representations},
  year={2014}
}

@inproceedings{jiang2026anomagic,
  title={Anomagic: Crossmodal prompt-driven zero-shot anomaly generation},
  author={Jiang, Yuxin and Luo, Wei and Zhang, Hui and Chen, Qiyu and Yao, Haiming and Shen, Weiming and Cao, Yunkang},
  booktitle={Proceedings of the AAAI Conference on Artificial Intelligence},
  volume={40},
  number={7},
  pages={5485--5493},
  year={2026}
}

@article{xu2025deltadeno,
  title={DeltaDeno: Zero-shot anomaly generation via delta-denoising attribution},
  author={Xu, Chaoran and Lv, Chengkan and Chen, Qiyu and Cao, Yunkang and Zhang, Feng and Zhang, Zhengtao},
  journal={arXiv preprint arXiv:2511.16920},
  year={2025}
}

\clearpage
\appendix
\normalsize
\setlength{\parindent}{0pt}
\section*{Appendix}
\label{sec:appendix_}

\setcounter{page}{1}
\setcounter{secnumdepth}{2}
\setcounter{section}{0}
\renewcommand{\thesection}{\Alph{section}}
\setcounter{table}{0}
\renewcommand{\thetable}{S\arabic{table}}
\setcounter{figure}{0}
\renewcommand{\thefigure}{S\arabic{figure}}

This appendix includes the following five parts:
\emph{(i)} We provide detailed descriptions of the datasets used in our experiments in Sec.~\ref{sec:_dataset_}.
\emph{(ii)} The introduction and implementation details of our CoPS method and the SOTA comparison methods are presented in Sec.~\ref{sec:_implement_}.
\emph{(iii)} Additional experimental results, including comparative experiments, ablation studies, and further analysis, are reported in Sec.~\ref{sec:_experiment_}.
\emph{(iv)} More detailed quantitative results are presented in Sec.~\ref{sec:_quantitative_}.
\emph{(v)} More detailed qualitative results are provided in Sec.~\ref{sec:_qualitative_}.

\section{Detailed Dataset Descriptions}
\label{sec:_dataset_}

\begin{table*}[t]\small
  \centering
  \caption{Key statistics of the 13 industrial and medical datasets used in our experiments. `\#' denotes the number of instances.}
    \begin{tabularx}{\textwidth}{p{1.1cm}|p{2.0cm}|>{\centering\arraybackslash}p{0.8cm}YYY>{\centering\arraybackslash}p{1.6cm}>{\centering\arraybackslash}p{1.5cm}}
    \noalign{\hrule height 0.4mm}
    Domain & Dataset & \#Class & \#Normal Image & \#Anomaly Image & Data Type & Pixel Label & Real-world \\
    \hline
    \multirow{5}{*}{Industrial} & MVTec-AD & 15    & 467   & 1258  & object \& texture & \cmark     & \cmark \\
          & VisA  & 12    & 962   & 1200  & object & \cmark     & \cmark \\
          & BTAD  & 3     & 451   & 290   & object \& texture & \cmark     & \cmark \\
          & MPDD  & 6     & 176   & 282   & object & \cmark     & \cmark \\
          & DTD-Synthetic & 12    & 357   & 947   & texture & \cmark     & \xmark \\
    \hline
    \multirow{8}{*}{Medical} & HeadCT & 1     & 100   & 100   & brain & \xmark     & \cmark \\
          & BrainMRI & 1     & 98    & 155   & brain & \xmark     & \cmark \\
          & Br35H & 1     & 1500  & 1500  & brain & \xmark     & \cmark \\
          & ISIC  & 1     & 0     & 379   & skin  & \cmark     & \cmark \\
          & CVC-ColonDB & 1     & 0     & 380   & colon & \cmark     & \cmark \\
          & CVC-ClinicDB & 1     & 0     & 612   & colon & \cmark     & \cmark \\
          & Kvasir & 1     & 0     & 1000  & colon & \cmark     & \cmark \\
          & Endo  & 1     & 0     & 200   & colon & \cmark     & \cmark \\
    \noalign{\hrule height 0.4mm}
    \end{tabularx}%
  \label{tab:dataset_}%
\end{table*}%

We evaluate the ZSAD performance of our proposed method on 13 publicly available datasets from industrial and medical domains.
As shown in Tab.~\ref{tab:dataset_}, we employ five industrial and eight medical datasets commonly used in prior studies.

\textbf{MVTec-AD}
\cite{bergmann2019mvtec} is one of the most challenging datasets in the industrial domain.
This dataset contains 15 high-resolution industrial product categories divided into texture and object groups, including over 70 types of defects.
In this work, we only use its labeled test set, which contains 467 normal and 1258 anomalous samples.

\textbf{VisA}
\cite{zou2022spot} is one of the largest datasets for industrial anomaly detection, including 10821 images across 12 categories of colored industrial parts.
The dataset covers diverse appearance defects under varying illumination and background conditions. 
In this work, we only use its labeled test set, which contains 962 normal samples and 1200 anomalous samples.

\textbf{BTAD}
\cite{mishra2021vt} is a real-world industrial anomaly detection dataset composed of 3 high-resolution categories.
It includes 451 normal and 290 anomalous samples with pixel-level annotations.
Similar to MVTec-AD, BTAD captures both structural and surface-level defects in practical inspection scenarios. 

\textbf{MPDD}
\cite{jezek2021deep} is a real-world industrial dataset focused on defect detection in metal parts.
The dataset captures fine-grained structural anomalies commonly encountered in industrial manufacturing.
In this work, we only use its labeled test set, which contains 176 normal samples and 282 anomalous samples.

\textbf{DTD-Synthetic}
\cite{aota2023zero} is a synthetic industrial dataset containing 12 texture categories with 357 normal and 947 anomalous images.
Despite being synthetically generated, it provides pixel-level anomaly annotations, enabling both image-level and pixel-level evaluation.

\textbf{HeadCT}
\cite{salehi2021multiresolution} is a medical anomaly detection dataset comprising head CT scans across 1 anatomical category with 100 normal and 100 anomalous images.
The dataset covers diverse pathological conditions and is widely used for evaluating anomaly detection methods in medical imaging.
Since HeadCT offers only image-level labels and lacks pixel-level annotations, it is primarily used for image-level evaluation.

\textbf{BrainMRI}
\cite{kanade2015brain} is a medical anomaly detection dataset consisting of brain MRI scans within a single anatomical class, containing 98 normal and 155 anomalous images.
The collection spans varied neuropathologies and serves as a standard benchmark for assessing medical anomaly detection methods.
As BrainMRI provides only image-level labels without pixel-level annotations, it is primarily used for image-level evaluation.

\textbf{Br35H}
\cite{hamada2020brain} is a medical anomaly detection dataset of brain MRI scans within a single anatomical class, comprising 1500 normal and 1500 anomalous images.
The images encompass a variety of brain pathologies and are widely adopted for benchmarking medical anomaly detection methods.
Also, Br35H provides only image-level labels without pixel-level annotations, so it is primarily used for image-level evaluation.

\textbf{ISIC}
\cite{codella2018skin} is a medical anomaly detection dataset of dermoscopic skin images within a single anatomical class, comprising 379 anomalous images and no normal images.
Each image is supplied with pixel-level lesion masks,
making the dataset a benchmark for evaluating pixel-level anomaly segmentation rather than image-level anomaly classification.

\textbf{CVC-ColonDB}
\cite{tajbakhsh2015automated} is a colonoscopy anomaly dataset containing 380 anomalous images and no normal images.
Every image is annotated with a pixel-level polyp mask,
establishing the dataset as a standard benchmark for evaluating pixel-level anomaly segmentation rather than image-level anomaly classification.

\textbf{CVC-ClinicDB}
\cite{bernal2015wm} is a colonoscopy anomaly dataset containing 612 anomalous images and no normal images, similar to CVC-ColonDB.
Each image includes a pixel-level polyp mask,
making the dataset a standard benchmark for evaluating pixel-level anomaly segmentation rather than image-level anomaly classification.

\textbf{Kvasir}
\cite{jha2019kvasir} is a colonoscopy anomaly dataset containing 1000 anomalous images and no normal images, similar to CVC-ColonDB.
Every image is annotated with a pixel-level polyp mask,
establishing the dataset as a standard benchmark for evaluating pixel-level anomaly segmentation rather than image-level anomaly classification.

\textbf{Endo}
\cite{hicks2021endotect} is a colonoscopy anomaly dataset containing 200 anomalous images and no normal images, similar to CVC-ColonDB.
Every image is annotated with a pixel-level polyp mask,
establishing the dataset as a standard benchmark for evaluating pixel-level anomaly segmentation rather than image-level anomaly classification.
\section{Additional Implementation Details}
\label{sec:_implement_}

This section provides additional implementation details of our method CoPS,
as well as descriptions and reproduction settings of other SOTA comparison methods. 
For a fair comparison, all methods are evaluated under the same CLIP backbone,
input resolution, and evaluation protocol.

\textbf{CoPS}
is our proposed method, which dynamically synthesizes visually conditioned prompts to adapt CLIP for zero-shot anomaly detection, achieving SOTA performance.
Following previous works \cite{zhou2024anomalyclip,cao2024adaclip,zhu2025fine},
we adopt the publicly available CLIP (ViT-L/14@336px) pre-trained by OpenAI~\cite{radford2021learning}. 
Input images are resized to \mbox{$518\times518$}.
Visual and textual embeddings are extracted from the final layers of the vision and text encoders, with a dimensionality of \mbox{$C=768$}.
In the 2nd to 9th layers of the text encoder, eight sets of four learnable tokens replace the input prefix to refine the textual representation.
All layers of the vision encoder employ both Q-KV and V-VV branches in parallel.
For ESTS, the context token length $K$, state token length $M$, and class token length $N$ are set to 6, 6, and 2, respectively.
The prototype extractor $\mathcal{P}_\theta$ is configured with 12 attention heads
and a two-layer feed-forward network whose hidden layer has the same dimensionality as the input.
For ICTS, the sampling count $R$ is set to 10.
The VAE employs two-layer MLPs for both the encoder $q_{\psi '}$ and decoder $p_{\psi ''}$, with hidden layers matching the input dimensionality.
For SAGA, the distance coefficient $\alpha$ and glocal coefficient $\beta$ are set to 0.3 and 0.9, respectively.
The default temperature hyperparameter $\tau$ is 0.07.
CoPS is trained using the Adam optimizer for 10 epochs with an initial learning rate of 0.001 and a batch size of 8.
During inference, a Gaussian filter with \mbox{$\sigma=4$} is applied to smooth the anomaly map.
The results are reported with the random seed fixed to 0 for reproducibility.
All experiments are conducted on a system equipped with a single NVIDIA GeForce RTX 3090 GPU and an Intel Xeon Gold 6226R CPU.

\textbf{WinCLIP}
\cite{jeong2023winclip} is the first work to employ frozen CLIP for ZSAD.  
It leverages window-based patch sampling and computes text-image similarity at the region level to localize anomalies.  
Anomaly scores are derived by aggregating the dissimilarity between visual patches and the textual description of normality.
This method does not require additional training data or fine-tuning, making it a training-free solution for ZSAD.
As the official implementation of WinCLIP is unavailable, we adopt the reproduced code from \cite{zhou2024anomalyclip}.

\textbf{APRIL-GAN}
\cite{chen2023april} builds on a frozen CLIP backbone and adds lightweight trainable linear layers
to project patch features into the shared image-text space for finer alignment with compositional prompts. 
It further maintains class-specific memory banks of normal references whose features are contrasted with test features to refine anomaly maps during inference.
These designs allow APRIL-GAN to perform zero-/few-shot anomaly classification and segmentation without task-specific retraining.
As APRIL-GAN adopts the same backbone (ViT-L/14@336px) and input resolution ($518\times518$) as ours,
we evaluate it directly using the official implementation and pre-trained weights.

\textbf{CLIP-AD}
\cite{chen2024clip} builds on APRIL-GAN's lightweight linear adapters and
further integrates representative vector selection and multi-scale feature fusion to produce both image-/pixel-level anomaly scores.
Since CLIP-AD originally uses an uncommon backbone (ViT-B/16@240px) and input resolution ($240\times240$),
we retrain it with the official code under our backbone (ViT-L/14@336px) and input resolution ($518\times518$).

\textbf{AdaCLIP}
\cite{cao2024adaclip} adapts CLIP for ZSAD by jointly fine-tuning
both vision and text encoders while learning hybrid prompts that combine globally optimized static tokens with per-image dynamic tokens.
The hybrid prompts guide the dual encoders to disentangle normal and abnormal semantics.
Since AdaCLIP adopts a non-standard evaluation protocol with multiple auxiliary datasets,
we retrain it with the official code under the same single-auxiliary setting as ours.

\begin{figure*}[t]
    \centering
    \includegraphics[width=\linewidth]{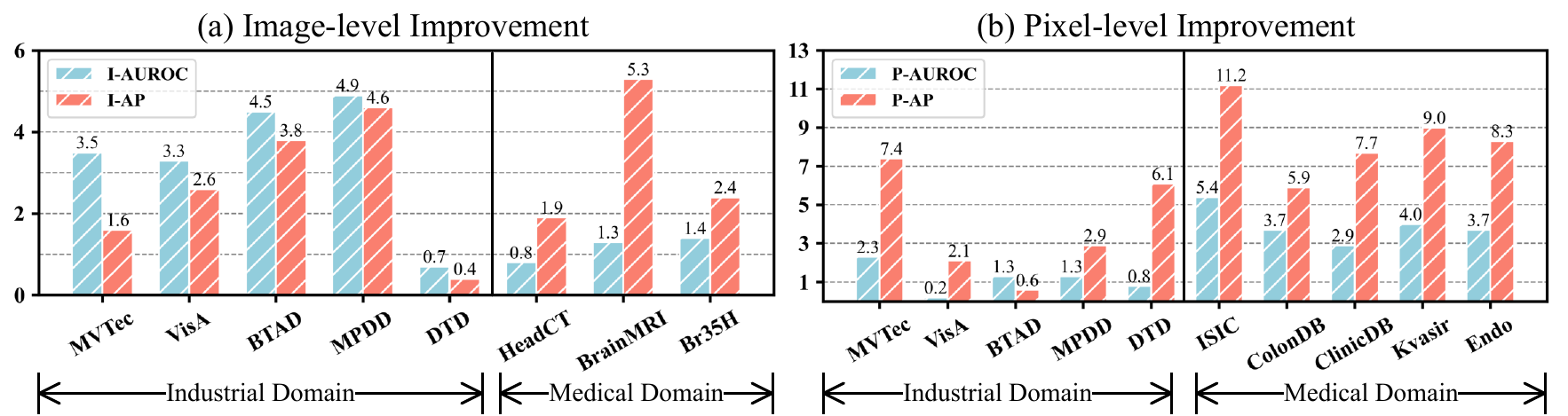}
    \caption{Performance improvements of CoPS over the baseline AnomalyCLIP across all 13 industrial and medical datasets.}
    \label{fig:improve_}
\end{figure*}

\textbf{AnomalyCLIP}
\cite{zhou2024anomalyclip} builds on CoOp \cite{zhou2022learning} by learning context tokens
for dual prompts that represent ``normal'' and ``anomalous'' states with frozen vision encoder and trainable text encoder.
Additionally, it employs consistent self-attention (i.e., V-VV) across the visual encoder layers
to emphasize diagonally prominent local features and improve fine-grained anomaly localization.
AnomalyCLIP serves as the baseline for our proposed method.
As AnomalyCLIP adopts the same backbone (ViT-L/14@336px) and input resolution ($518\times518$) as ours,
we evaluate it directly using the official implementation and pre-trained weights.

\textbf{FAPrompt}
\cite{zhu2025fine} introduces a fine-grained prompt tuning method for ZSAD which is built upon CoCoOp \cite{zhou2022conditional}.
It synthesizes compound abnormality prompts by leveraging the mapped top-K abnormality-prior patch features.
This approach allows for more flexible and adaptive prompt design, improving the model's ability to capture diverse anomalies.
Although FAPrompt provides official code, no pre-trained weights are available.
Thus, we retrain it with the official implementation under the same evaluation protocol.

\section{Extended Experimental Analysis}
\label{sec:_experiment_}

We provide additional comparative and ablation experiments to further validate the effectiveness of CoPS.

\textbf{Performance improvement.}
Fig.~\ref{fig:improve_} illustrates the relative improvements of CoPS over the prompt-tuning baseline, AnomalyCLIP.
CoPS achieves consistent gains over AnomalyCLIP on all 13 datasets,
particularly showing larger improvements in image-level performance (Fig.~\ref{fig:improve_}(a)) for industrial datasets
and in pixel-level performance (Fig.~\ref{fig:improve_}(b)) for medical datasets.
These results demonstrate the effectiveness of CoPS's prototype extraction, class sampling, and glocal alignment components.

\begin{table*}[t]\small
  \centering
  \caption{Comparison of accuracy and efficiency across various methods on MVTec-AD dataset.
  The best results are highlighted in bold.}
    \begin{tabularx}{\textwidth}{p{1.8cm}|ZYZY|ZZZY}
    \noalign{\hrule height 0.4mm}
    Metric~$\rightarrow$ & I-AUROC & I-AP  & P-AUROC & P-AP  & Model Size & Train Mem. & Test Mem. & 1/FPS \\
    Method~$\downarrow$ & (\%)  & (\%)  & (\%)  & (\%)  & (MB)  & (GB)  & (GB)  & (ms) \\
    \hline
    WinCLIP & 91.8  & 95.1  & 85.1  & 18.0  & - & - & \textbf{2.0} & 840 \\
    APRIL-GAN & 86.1  & 93.5  & 87.6  & 40.8  & 12    & \textbf{5.5}   & 3.3   & \textbf{105} \\
    CLIP-AD & 89.8  & 95.3  & 89.8  & 40.0  & \textbf{8.7}   & 6.7   & 3.4   & 115 \\
    AdaCLIP & 92.0  & 96.4  & 86.8  & 38.1  & 41    & 10  & 3.3   & 183 \\
    AnomalyCLIP & 91.5  & 96.2  & 91.1  & 34.5  & 22    & 6.9   & 2.7   & 131 \\
    FAPrompt & 91.9  & 95.9  & 90.4  & 34.6  & 39    & 11    & 2.7   & 174 \\
    \hline
    CoPS  & \textbf{95.0} & \textbf{97.8} & \textbf{93.4} & \textbf{41.9} & 19    & 7.1   & 2.7   & 168 \\
    \noalign{\hrule height 0.4mm}
    \end{tabularx}%
  \label{tab:fps_}%
\end{table*}%

\textbf{Computational efficiency.}
Tab.~\ref{tab:fps_} compares the accuracy and computational efficiency of various methods
in terms of image-level and pixel-level performance, model size, memory consumption, and inference speed.
CoPS achieves the best performance across all four evaluation metrics (I-AUROC, I-AP, P-AUROC, and P-AP), outperforming all prior methods.
In terms of efficiency, CoPS maintains a competitive model size (19 MB) and moderate memory usage (7.1 GB for training and 2.7 GB for testing),
while achieving an inference speed of 168 ms per frame.
Although WinCLIP requires no fine-tuning, it suffers from high inference latency.
APRIL-GAN and CLIP-AD offer relatively efficient inference, but their accuracy remains notably limited.
Furthermore, CoPS outperforms AdaCLIP in both accuracy and computational efficiency.
While its inference time is marginally higher than that of AnomalyCLIP,
CoPS yields significantly improved results on both image-level and pixel-level metrics.
Finally, CoPS also outperforms the SOTA method FAPrompt, achieving consistent improvements across all evaluation metrics.
These results demonstrate that CoPS offers a favorable balance between accuracy and efficiency,
making it a practical and scalable solution for real-world ZSAD tasks.

\begin{table*}[t]\small
  \centering
  \caption{Performance ablation of different backbones and input image sizes, as measured by I-AUROC\%, I-AP\%, P-AUROC\%, and P-AP\%.
  The best results are highlighted in bold, and the second-best results are underlined.}
    \begin{tabularx}{\textwidth}{YY|YYYY}
    \noalign{\hrule height 0.4mm}
    Backbone &  Image Size & I-AUROC & I-AP  & P-AUROC & P-AP \\
    \hline
    ViT-B/16@224px & $336\times336$ & 85.4  & 93.4  & 88.6  & 31.2 \\
    ViT-L/14@224px & $336\times336$ & 89.9  & 95.1  & 91.5  & 34.8 \\
    ViT-L/14@224px & $518\times518$ & 90.0  & 95.4  & 91.8  & 40.6 \\
    ViT-L/14@336px & $336\times336$ & 92.2  & 96.4  & \underline{92.6}  & 37.1 \\
    ViT-L/14@336px & $518\times518$ & \textbf{95.0} & \textbf{97.8} & \textbf{93.4} & \textbf{41.9} \\
    ViT-L/14@336px & $700\times700$ & \underline{92.4}  & \underline{97.2}  & 92.5  & \underline{41.8} \\
    \noalign{\hrule height 0.4mm}
    \end{tabularx}%
  \label{tab:backbone_}%
\end{table*}%

\begin{figure}[t]
    \centering
    \includegraphics[width=0.7\linewidth]{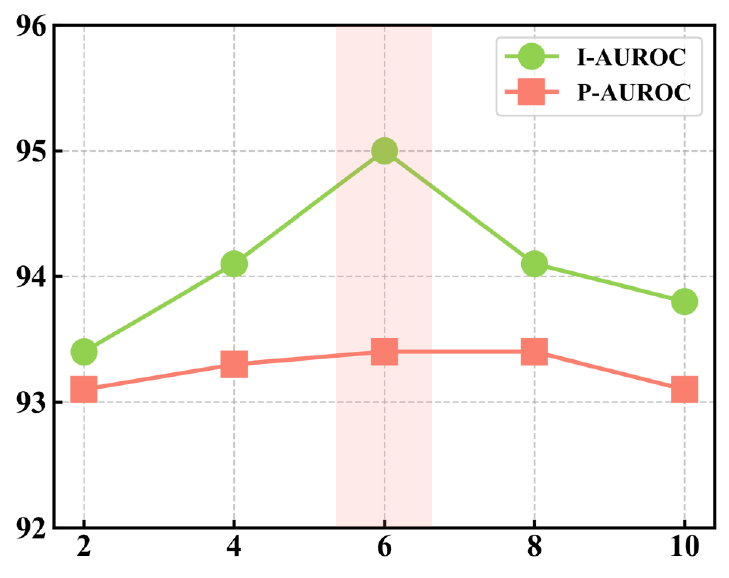}
    \caption{Performance ablation of context token length $K$.}
    \label{fig:K_}
\end{figure}

\textbf{Influence of pre-trained backbone.}
As shown in Tab.~\ref{tab:backbone_},
we analyze the impact of different pre-trained CLIP backbones and input image resolutions on model performance.
The results indicate a consistent performance improvement with larger backbones and higher input resolutions.
However, when the resolution increases to \mbox{$700 \times 700$}, performance declines due to semantic misalignment caused by overly small patch sizes.
The best performance is achieved using ViT-L/14@336px with an input size of \mbox{$518 \times 518$},
which is also adopted as the default configuration in our experiments.
These settings are widely adopted in most existing methods~\cite{chen2023april,zhou2024anomalyclip,cao2024adaclip}.

\textbf{Influence of context token length.}
As illustrated in Fig.~\ref{fig:K_},
increasing the context token length $K$ initially improves both I-AUROC and P-AUROC, reaching peak performance at $K = 6$.  
Further increasing $K$ beyond this point leads to performance degradation, likely due to overfitting in the prompt space.  
Therefore, we adopt $K = 6$ as the default setting in all experiments.
\section{More Quantitative Results}
\label{sec:_quantitative_}

As the industrial datasets contain multiple categories,
we report the detailed performance of CoPS across all categories and further compare it with all SOTA methods on MVTec-AD and VisA.
Specifically, Tabs. \ref{tab:mvtec_i_} and \ref{tab:mvtec_p_}
present the image-level and pixel-level performance of SOTA methods on MVTec-AD across four metrics for each category.
These metrics include image-level AUROC (I-AUROC) and image-level AP (I-AP) for classification,
and pixel-level AUROC (P-AUROC) and pixel-level AP (P-AP) for segmentation.
Tabs. \ref{tab:visa_i_} and \ref{tab:visa_p_}
report the image-level and pixel-level performance of SOTA methods on VisA, also evaluated using four metrics per category.
Tabs. \ref{tab:dtd_}--\ref{tab:mpdd_}
show the per-category results of CoPS on DTD-Synthetic, BTAD, and MPDD, respectively.

\begin{table}[t]\small
  \centering
  \caption{Performance of CoPS on each category of the DTD-Synthetic dataset under the ZSAD setting.}
  \begin{tabularx}{0.48\textwidth}{p{1.8cm}|Y>{\centering\arraybackslash}p{0.8cm}Y>{\centering\arraybackslash}p{0.8cm}}
    \noalign{\hrule height 0.4mm}
    Category & I-AUROC & I-AP  & P-AUROC & P-AP \\
    \hline
    Woven\_001 & 100 & 100 & 99.5 & 63.3 \\
    Woven\_127 & 92.5 & 94.2 & 96.0 & 42.9 \\
    Woven\_104 & 99.1 & 99.8 & 97.5 & 60.4 \\
    Stratified\_154 & 98.9 & 99.8 & 99.7 & 79.4 \\
    Blotchy\_099 & 99.2 & 99.8 & 99.2 & 66.8 \\
    Woven\_068 & 96.4 & 98.0 & 98.3 & 37.6 \\
    Woven\_125 & 100 & 100 & 99.4 & 66.8 \\
    Marbled\_078 & 98.6 & 99.7 & 99.1 & 61.1 \\
    Perforated\_037 & 93.1 & 98.3 & 97.1 & 55.9 \\
    Mesh\_114 & 85.0 & 93.9 & 97.5 & 49.9 \\
    Fibrous\_183 & 99.4 & 99.9 & 99.2 & 67.1 \\
    Matted\_069 & 80.3 & 94.4 & 98.5 & 50.8 \\
    \hline
    Average & 95.2 & 98.1 & 98.4 & 58.5 \\
    \noalign{\hrule height 0.4mm}
  \end{tabularx}
  \label{tab:dtd_}
\end{table}

\begin{table*}[t]
  \centering
  \small
  \begin{minipage}[t]{0.48\textwidth}
    \centering
    \captionof{table}{Performance of CoPS on each category of the BTAD dataset under the ZSAD setting.}
    \begin{tabularx}{\textwidth}{p{1.8cm}|Y>{\centering\arraybackslash}p{0.8cm}Y>{\centering\arraybackslash}p{0.8cm}}
      \noalign{\hrule height 0.4mm}
      Category & I-AUROC & I-AP  & P-AUROC & P-AP \\
      \hline
      01 & 95.8 & 98.4 & 93.7 & 46.5 \\
      02 & 87.3 & 97.9 & 95.2 & 61.7 \\
      03 & 97.7 & 88.2 & 95.0 & 19.5 \\
      \hline
      Average & 93.6 & 94.9 & 94.6 & 42.6 \\
      \noalign{\hrule height 0.4mm}
    \end{tabularx}
    \label{tab:btad_}
    \end{minipage}
    \hfil
    \begin{minipage}[t]{0.48\textwidth}
    \centering
    \captionof{table}{Performance of CoPS on each category of the MPDD dataset under the ZSAD setting.}
    \begin{tabularx}{\textwidth}{p{1.8cm}|Y>{\centering\arraybackslash}p{0.8cm}Y>{\centering\arraybackslash}p{0.8cm}}
      \noalign{\hrule height 0.4mm}
      Category & I-AUROC & I-AP  & P-AUROC & P-AP \\
      \hline
      bracket\_black & 63.4 & 75.1 & 97.3 & 6.70 \\
      bracket\_brown & 60.6 & 76.4 & 95.8 & 12.4 \\
      bracket\_white & 77.9 & 66.9 & 99.0 & 6.92 \\
      connector & 80.0 & 72.1 & 97.1 & 21.1 \\
      metal\_plate & 92.7 & 97.4 & 98.3 & 87.6 \\
      tubes & 97.2 & 98.7 & 97.6 & 50.9 \\
      \hline
      Average & 78.6 & 81.1 & 97.5 & 30.9 \\
      \noalign{\hrule height 0.4mm}
    \end{tabularx}
    \label{tab:mpdd_}
  \end{minipage}
\end{table*}


\begin{table*}[t]\small
  \centering
  \caption{Performance comparison of various SOTA methods on each category of the MVTec-AD dataset under ZSAD setting, as measured by I-AUROC\% / I-AP\%.
   The best results are highlighted in bold, and the second-best results are underlined.}
    \begin{tabularx}{\textwidth}{p{1.5cm}|YYYY|YY|Y}
    \noalign{\hrule height 0.4mm}
    Method~$\rightarrow$ & \multicolumn{4}{c|}{Prompt Design} & \multicolumn{3}{c}{Prompt Learning} \\
    \hline
    Category~$\downarrow$ & WinCLIP & APRIL-GAN & CLIP-AD & AdaCLIP & AnomalyCLIP & FAPrompt & \textbf{CoPS} \\
    \hline
    bottle & \textbf{99.2} / 98.3 & 92.0 / 97.7 & \underline{96.4} / \textbf{98.8} & 95.6 / \underline{98.6} & 88.7 / 96.8 & 89.3 / 96.4 & 92.5 / 97.8 \\
    cable & 86.5 / 86.2 & \underline{88.2} / \underline{92.9} & 80.4 / 88.9 & 79.0 / 87.3 & 70.3 / 81.7 & 73.3 / 81.8 & \textbf{89.6} / \textbf{94.1} \\
    capsule & 72.9 / 93.4 & 79.8 / 95.4 & 82.8 / 96.4 & 89.3 / 97.8 & 89.5 / 97.8 & \underline{93.0} / \underline{98.6} & \textbf{95.5} / \textbf{99.1} \\
    carpet & \textbf{100} / \underline{99.9} & 99.4 / 99.8 & 99.5 / 99.8 & \textbf{100} / \textbf{100} & \underline{99.9} / \underline{99.9} & \textbf{100} / \textbf{100} & \textbf{100} / \textbf{100} \\
    grid & 98.8 / \textbf{99.8} & 86.2 / 94.9 & 94.1 / 97.9 & \underline{99.2} / \underline{99.7} & 97.8 / 99.3 & 98.2 / 99.3 & \textbf{99.3} / \textbf{99.8} \\
    hazelnut & 93.9 / 96.3 & 89.4 / 94.6 & \underline{98.0} / \underline{99.0} & 95.5 / 97.5 & 97.2 / 98.5 & 96.8 / 98.2 & \textbf{98.8} / \textbf{99.3} \\
    leather & \textbf{100} / \textbf{100} & 99.7 / \underline{99.9} & \textbf{100} / \textbf{100} & \textbf{100} / \textbf{100} & 99.8 / \underline{99.9} & \underline{99.9} / \textbf{100} & \textbf{100} / \textbf{100} \\
    metal\_nut & \textbf{97.1} / \underline{97.9} & 68.2 / 91.8 & 75.1 / 94.4 & 79.9 / 95.6 & \underline{92.4} / \textbf{98.1} & 87.6 / 97.1 & 89.6 / 97.5 \\
    pill & 79.1 / 96.5 & 80.8 / 96.1 & 87.7 / 97.6 & \textbf{92.6} / \textbf{98.6} & 81.1 / 95.3 & 89.4 / 97.9 & \underline{92.0} / \underline{98.2} \\
    screw & 83.3 / 88.4 & 85.1 / 93.6 & \textbf{89.1} / \textbf{96.2} & 83.9 / 93.0 & 82.1 / 92.9 & \underline{86.9} / \underline{94.7} & 84.2 / 94.2 \\
    tile & \textbf{100} / \underline{99.9} & 99.8 / \underline{99.9} & 99.6 / 99.8 & 99.7 / \underline{99.9} & \textbf{100} / \textbf{100} & 99.8 / \underline{99.9} & \underline{99.9} / \textbf{100} \\
    toothbrush & 87.5 / 96.7 & 53.2 / 71.9 & 76.1 / 90.2 & \underline{95.2} / \underline{97.9} & 85.3 / 93.9 & 83.9 / 92.7 & \textbf{96.1} / \textbf{98.8} \\
    transistor & 88.0 / 74.9 & 80.9 / 77.6 & 79.3 / 73.7 & 82.0 / 83.8 & \textbf{93.9} / \textbf{92.1} & 85.1 / 83.1 & \underline{90.9} / \underline{88.9} \\
    wood & \textbf{99.4} / 98.8 & \underline{98.9} / \textbf{99.6} & \underline{98.9} / \textbf{99.6} & 98.5 / \underline{99.5} & 96.9 / 99.2 & 97.8 / 99.4 & 98.2 / \underline{99.5} \\
    zipper & 91.5 / 98.9 & 89.4 / 97.1 & 88.6 / 96.9 & 89.4 / 97.1 & \textbf{98.4} / \textbf{99.5} & 97.4 / 99.3 & \underline{97.8} / \underline{99.4} \\
    \hline
    Average & 91.8 / 95.1 & 86.1 / 93.5 & 89.8 / 95.3 & \underline{92.0} / \underline{96.4} & 91.5 / 96.2 & 91.9 / 95.9 & \textbf{95.0} / \textbf{97.8} \\
    \noalign{\hrule height 0.4mm}
    \end{tabularx}%
  \label{tab:mvtec_i_}%
\end{table*}

\begin{table*}[t]\small
  \centering
  \caption{Performance comparison of various SOTA methods on each category of the MVTec-AD dataset under ZSAD setting, as measured by P-AUROC\% / P-AP\%.
   The best results are highlighted in bold, and the second-best results are underlined.}
    \begin{tabularx}{\textwidth}{p{1.5cm}|YYYY|YY|Y}
    \noalign{\hrule height 0.4mm}
    Method~$\rightarrow$ & \multicolumn{4}{c|}{Prompt Design} & \multicolumn{3}{c}{Prompt Learning} \\
    \hline
    Category~$\downarrow$ & WinCLIP & APRIL-GAN & CLIP-AD & AdaCLIP & AnomalyCLIP & FAPrompt & \textbf{CoPS} \\
    \hline
    bottle      & 89.5 / 49.8 & 83.5 / 53.0 & \underline{91.2} / 56.8 & 83.8 / 49.8 & 90.4 / 55.3 & 90.5 / \underline{57.4} & \textbf{92.8} / \textbf{61.8} \\
    cable       & 77.0 / 6.20 & 72.2 / \textbf{18.2} & 76.2 / \underline{17.3} & \textbf{85.6} / 16.5 & 78.9 / 12.3 & 78.0 / 11.9 & \underline{79.0} / 15.1 \\
    capsule     & 86.9 / 8.60 & 92.0 / \underline{29.6} & 95.1 / 27.2 & 86.2 / 24.8 & \underline{95.8} / 27.7 & 95.4 / 26.0 & \textbf{97.4} / \textbf{30.6} \\
    carpet      & 95.4 / 25.9 & 98.4 / \underline{67.5} & \underline{99.1} / 65.4 & 94.8 / 63.5 & 98.8 / 56.6 & 99.0 / 60.7 & \textbf{99.3} / \textbf{74.5} \\
    grid        & 82.2 / 5.70 & 95.8 / \textbf{36.5} & 96.3 / \underline{30.7} & 90.6 / 27.8 & \underline{97.3} / 24.1 & 97.2 / 23.9 & \textbf{97.8} / 26.9 \\
    hazelnut    & 94.3 / 33.3 & 96.1 / 49.7 & 97.2 / \underline{59.2} & \textbf{98.7} / \textbf{69.5} & 97.2 / 43.4 & \underline{97.5} / 45.9 & 97.4 / 51.1 \\
    leather     & 96.7 / 20.4 & 99.1 / \underline{52.3} & \textbf{99.3} / 50.5 & 97.8 / \textbf{53.6} & 98.6 / 22.7 & 98.6 / 25.1 & \underline{99.2} / 36.0 \\
    metal\_nut  & 61.0 / 10.8 & 65.5 / 25.9 & 58.9 / 21.2 & 55.4 / 19.9 & \underline{74.6} / \underline{26.4} & 68.0 / 23.3 & \textbf{88.5} / \textbf{39.0} \\
    pill        & 80.0 / 7.00 & 76.2 / 23.6 & 83.7 / 26.1 & 77.5 / 25.8 & \underline{91.8} / \underline{34.1} & 90.2 / 31.2 & \textbf{92.3} / \textbf{35.0} \\
    screw       & 89.6 / 5.40 & 97.8 / 33.7 & \underline{98.7} / \underline{39.1} & \textbf{99.2} / \textbf{41.6} & 97.5 / 27.5 & 97.3 / 24.7 & 98.2 / 22.4 \\
    tile        & 77.6 / 21.2 & 92.7 / \underline{66.3} & 94.5 / 65.2 & 83.9 / 48.8 & 94.7 / 61.7 & \underline{96.2} / 64.6 & \textbf{97.9} / \textbf{79.2} \\
    toothbrush  & 86.9 / 5.50 & \textbf{95.8} / \textbf{43.2} & 92.7 / \underline{29.9} & 93.4 / 24.7 & 91.9 / 19.3 & 89.8 / 15.9 & \underline{94.9} / 25.0 \\
    transistor  & \underline{74.7} / \textbf{20.2} & 62.4 / 11.7 & \textbf{75.5} / 14.2 & 71.4 / 11.9 & 70.8 / 15.6 & 70.8 / 16.1 & 73.6 / \underline{17.0} \\
    wood        & 93.4 / 32.9 & 95.8 / \underline{61.8} & \underline{96.9} / 59.4 & 91.2 / 56.6 & 96.4 / 52.6 & \underline{96.9} / 55.2 & \textbf{97.5} / \textbf{68.4} \\
    zipper      & 91.6 / 19.4 & 91.1 / \underline{38.7} & \underline{92.8} / 38.5 & 91.8 / 36.0 & 91.2 / \underline{38.7} & 90.7 / 37.7 & \textbf{95.2} / \textbf{46.3} \\
    \hline
    Average     & 85.1 / 18.0 & 87.6 / \underline{40.8} & 89.8 / 40.0 & 86.8 / 38.1 & \underline{91.1} / 34.5 & 90.4 / 34.6 & \textbf{93.4} / \textbf{41.9} \\
    \noalign{\hrule height 0.4mm}
    \end{tabularx}%
  \label{tab:mvtec_p_}%
\end{table*}


\begin{table*}[t]\small
  \centering
    \caption{Performance comparison of various SOTA methods on each category of the VisA dataset under ZSAD setting, as measured by I-AUROC\% / I-AP\%.
   The best results are highlighted in bold, and the second-best results are underlined.}
    \begin{tabularx}{\textwidth}{p{1.5cm}|YYYY|YY|Y}
    \noalign{\hrule height 0.4mm}
    Method~$\rightarrow$ & \multicolumn{4}{c|}{Prompt Design} & \multicolumn{3}{c}{Prompt Learning} \\
    \hline
    Category~$\downarrow$ & WinCLIP & APRIL-GAN & CLIP-AD & AdaCLIP & AnomalyCLIP & FAPrompt & \textbf{CoPS} \\
    \hline
    candle     & \underline{95.4} / \underline{95.6} & 82.5 / 85.9 & 89.4 / 91.6 & \textbf{95.9} / \textbf{96.4} & 80.9 / 82.6 & 84.5 / 87.1 & 87.8 / 91.0 \\
    capsules   & 85.0 / 80.9 & 62.3 / 74.6 & 75.2 / 86.6 & 81.1 / 86.7 & 82.7 / 89.4 & \textbf{91.1} / \textbf{95.8} & \underline{88.9} / \underline{93.4} \\
    cashew     & \textbf{92.1} / \underline{95.2} & 86.7 / 93.9 & 83.7 / 92.4 & \underline{89.6} / \textbf{95.4} & 76.0 / 89.3 & 89.2 / \textbf{95.4} & 87.1 / 94.6 \\
    chewinggum & 96.5 / 98.8 & 96.5 / 98.4 & 95.6 / 98.1 & \textbf{98.5} / \textbf{99.4} & 97.2 / 98.8 & 97.4 / 99.0 & \underline{98.1} / \underline{99.2} \\
    fryum      & 80.3 / 92.5 & \underline{93.8} / 97.0 & 78.7 / 90.4 & 89.5 / 95.1 & 92.7 / 96.6 & \textbf{96.0} / \textbf{98.2} & \underline{93.8} / \underline{97.4} \\
    macaroni1  & 76.2 / 64.5 & 69.5 / 67.5 & 80.0 / 81.1 & \underline{86.3} / 85.0 & \textbf{86.7} / \underline{85.5} & 81.2 / 81.5 & 84.1 / \textbf{85.7} \\
    macaroni2  & 63.7 / 65.2 & 65.7 / 64.9 & 67.0 / 65.3 & 56.7 / 54.3 & \underline{72.2} / \underline{70.8} & \textbf{72.5} / \textbf{72.6} & 70.5 / 69.3 \\
    pcb1       & 73.6 / 74.6 & 50.6 / 54.6 & 68.6 / 72.5 & 74.0 / 73.5 & \underline{85.2} / \underline{86.7} & 69.4 / 74.1 & \textbf{86.6} / \textbf{89.1} \\
    pcb2       & 51.2 / 44.2 & \textbf{71.6} / \textbf{73.8} & 69.7 / 71.4 & \underline{71.1} / \underline{71.6} & 62.0 / 64.4 & 66.2 / 67.8 & 67.1 / 69.1 \\
    pcb3       & \underline{73.4} / 66.2 & 66.9 / 70.5 & 67.3 / 71.9 & \textbf{75.2} / \textbf{77.9} & 61.7 / 69.4 & 68.2 / \underline{75.5} & 66.4 / 71.3 \\
    pcb4       & 79.6 / 70.1 & 94.6 / 94.8 & \underline{96.2} / \underline{96.0} & 89.6 / 89.8 & 93.9 / 94.3 & 95.3 / 95.1 & \textbf{97.7} / \textbf{97.3} \\
    pipe\_fryum& 69.7 / 82.1 & 89.4 / 94.6 & 86.5 / 93.7 & 88.8 / 93.9 & 92.3 / 96.3 & \textbf{97.2} / \underline{98.6} & \underline{97.1} / \textbf{98.7} \\
    \hline
    Average    & 78.1 / 77.5 & 78.0 / 81.4 & 79.8 / 84.3 & 83.0 / 84.9 & 82.1 / 85.4 & \underline{84.0} / \underline{86.7} & \textbf{85.4} / \textbf{88.0} \\
    \noalign{\hrule height 0.4mm}
    \end{tabularx}%
  \label{tab:visa_i_}%
\end{table*}

\begin{table*}[t]\small
  \centering
  \caption{Performance comparison of various SOTA methods on each category of the VisA dataset under ZSAD setting, as measured by P-AUROC\% / P-AP\%.
  The best results are highlighted in bold, and the second-best results are underlined.}
  \begin{tabularx}{\textwidth}{p{1.5cm}|YYYY|YY|Y}
  \noalign{\hrule height 0.4mm}
  Method~$\rightarrow$ & \multicolumn{4}{c|}{Prompt Design} & \multicolumn{3}{c}{Prompt Learning} \\
  \hline
  Category~$\downarrow$ & WinCLIP & APRIL-GAN & CLIP-AD & AdaCLIP & AnomalyCLIP & FAPrompt & \textbf{CoPS} \\
  \hline
  candle      & 88.9 / 2.40 & 97.8 / 29.9 & \underline{98.7} / \underline{36.6} & 98.6 / \textbf{45.3} & \textbf{98.8} / 25.6 & \textbf{98.8} / 25.4 & 98.2 / 25.9 \\
  capsules    & 81.6 / 1.40 & \textbf{97.5} / \textbf{40.0} & \underline{97.4} / \underline{38.5} & 96.1 / 18.2 & 94.9 / 29.3 & 96.4 / 30.9 & 95.6 / 31.3 \\
  cashew      & 84.7 / 4.80 & 86.0 / 15.1 & 91.4 / 24.1 & \textbf{97.2} / \textbf{44.8} & 93.7 / 19.6 & 93.8 / 17.6 & \underline{95.5} / \underline{25.1} \\
  chewinggum  & 93.3 / 24.0 & \textbf{99.5} / \underline{83.6} & 99.2 / 83.4 & 99.2 / \textbf{87.6} & 99.2 / 56.3 & \underline{99.4} / 61.3 & \textbf{99.5} / 65.5 \\
  fryum       & 88.5 / 11.1 & 92.0 / 22.1 & 93.0 / 22.4 & 93.6 / \underline{24.0} & \underline{94.6} / 22.6 & 94.1 / 21.4 & \textbf{94.7} / \textbf{26.1} \\
  macaroni1   & 70.9 / 0.03 & \textbf{98.8} / \underline{24.8} & \underline{98.7} / 23.2 & \textbf{98.8} / \textbf{27.1} & 98.3 / 14.9 & 98.1 / 12.9 & 98.5 / 12.8 \\
  macaroni2   & 59.3 / 0.02 & \underline{97.8} / \textbf{6.80} & 97.6 / 2.30 & \textbf{98.2} / \underline{3.00} & 97.6 / 1.50 & 96.5 / 0.88 & 96.7 / 1.69 \\
  pcb1        & 61.2 / 0.40 & 92.7 / 8.40 & 92.6 / 7.20 & 90.7 / 7.80 & \underline{94.0} / 8.60 & \textbf{95.6} / \underline{9.72} & 93.7 / \textbf{9.76} \\
  pcb2        & 71.6 / 0.40 & 89.8 / \underline{15.4} & 91.0 / 8.20 & 91.3 / \textbf{17.5} & \underline{92.4} / 9.10 & 92.2 / 8.62 & \textbf{92.7} / 8.18 \\
  pcb3        & 85.3 / 0.70 & \underline{88.4} / \underline{14.1} & 87.5 / 11.7 & 87.7 / \textbf{16.1} & 88.3 / 4.30 & 88.0 / 3.66 & \textbf{89.8} / 5.71 \\
  pcb4        & 94.4 / 15.5 & 94.6 / 24.9 & \underline{95.9} / 31.2 & 94.6 / 34.2 & 95.7 / 30.6 & \textbf{97.2} / \textbf{38.3} & \underline{95.9} / \underline{35.1} \\
  pipe\_fryum & 75.4 / 4.40 & 96.0 / 23.6 & 96.9 / 27.2 & 95.7 / 24.4 & \textbf{98.2} / \underline{33.2} & \underline{97.8} / 26.3 & \textbf{98.2} / \textbf{33.3} \\
  \hline
  Average     & 79.6 / 5.00 & 94.2 / 25.7 & 95.0 / \underline{26.3} & 95.1 / \textbf{29.2} & \underline{95.5} / 21.3 & \textbf{95.7} / 21.4 & \textbf{95.7} / 23.4 \\
  \noalign{\hrule height 0.4mm}
  \end{tabularx}%
  \label{tab:visa_p_}
\end{table*}
\section{More Qualitative Results}
\label{sec:_qualitative_}

\begin{figure*}[t]
    \centering
    \includegraphics[width=\linewidth]{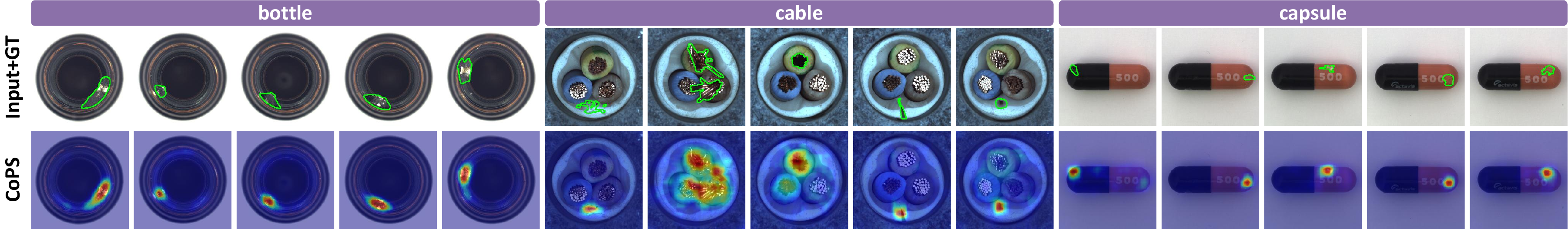}
    \caption{Qualitative segmentation results for the bottle, cable, and capsule categories from the MVTec-AD dataset.}
    \label{fig:mvtec1_}
\end{figure*}

\begin{figure*}[t]
    \centering
    \includegraphics[width=\linewidth]{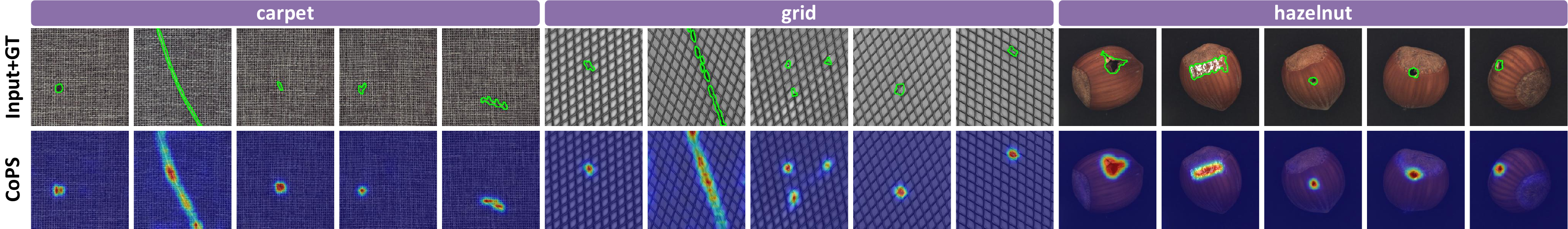}
    \caption{Qualitative segmentation results for the carpet, grid, and hazelnut categories from the MVTec-AD dataset.}
    \label{fig:mvtec2_}
\end{figure*}

\begin{figure*}[t]
    \centering
    \includegraphics[width=\linewidth]{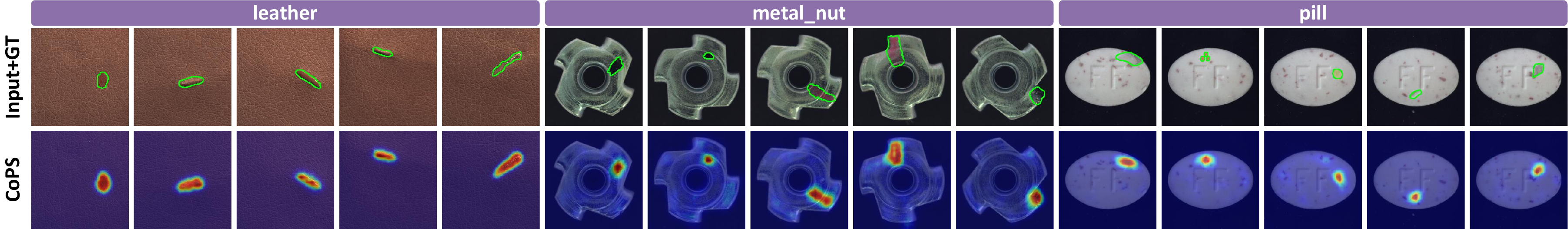}
    \caption{Qualitative segmentation results for the leather, metal nut, and pill categories from the MVTec-AD dataset.}
    \label{fig:mvtec3_}
\end{figure*}

\begin{figure*}[t]
    \centering
    \includegraphics[width=\linewidth]{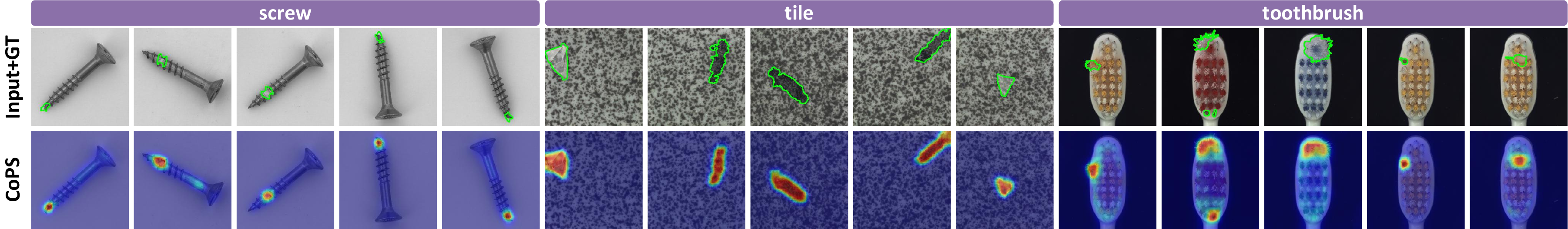}
    \caption{Qualitative segmentation results for the screw, tile, and toothbrush categories from the MVTec-AD dataset.}
    \label{fig:mvtec4_}
\end{figure*}

\begin{figure*}[t]
    \centering
    \includegraphics[width=\linewidth]{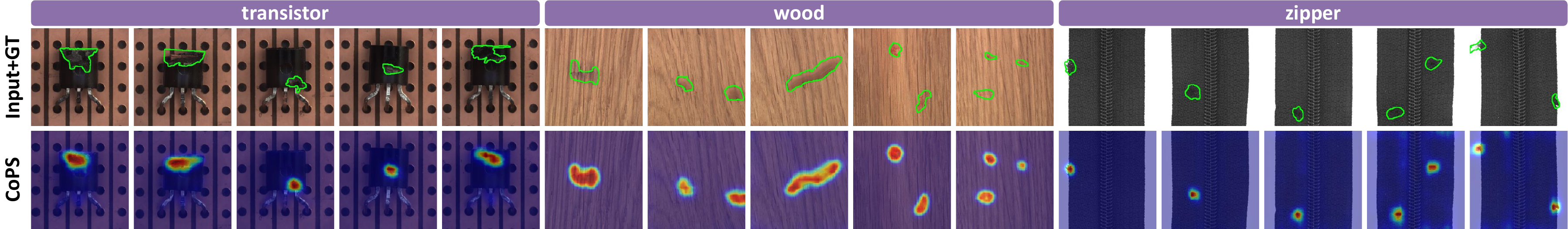}
    \caption{Qualitative segmentation results for the transistor, wood, and zipper categories from the MVTec-AD dataset.}
    \label{fig:mvtec5_}
\end{figure*}

\begin{figure*}[t]
    \centering
    \includegraphics[width=\linewidth]{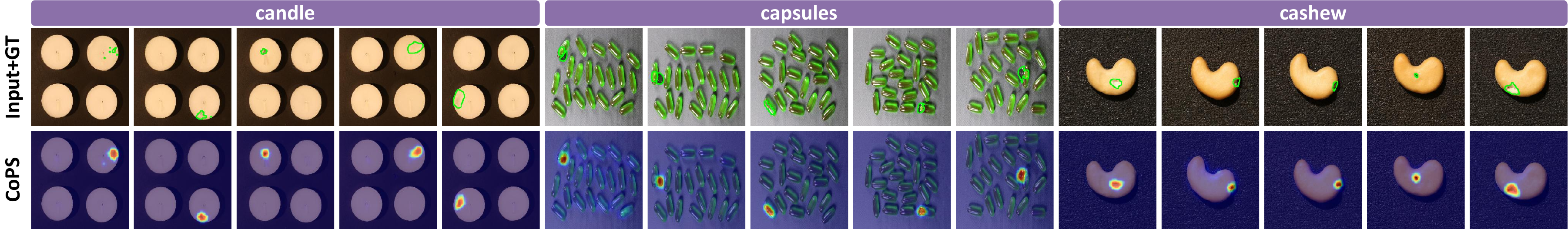}
    \caption{Qualitative segmentation results for the candle, capsules, and cashew categories from the VisA dataset.}
    \label{fig:visa1_}
\end{figure*}

\begin{figure*}[t]
    \centering
    \includegraphics[width=\linewidth]{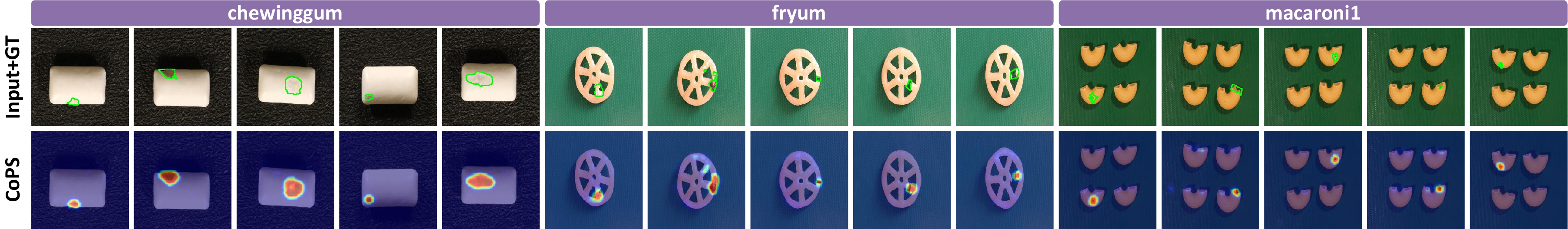}
    \caption{Qualitative segmentation results for the chewinggum, fryum, and macaroni1 categories from the VisA dataset.}
    \label{fig:visa2_}
\end{figure*}

\begin{figure*}[t]
    \centering
    \includegraphics[width=\linewidth]{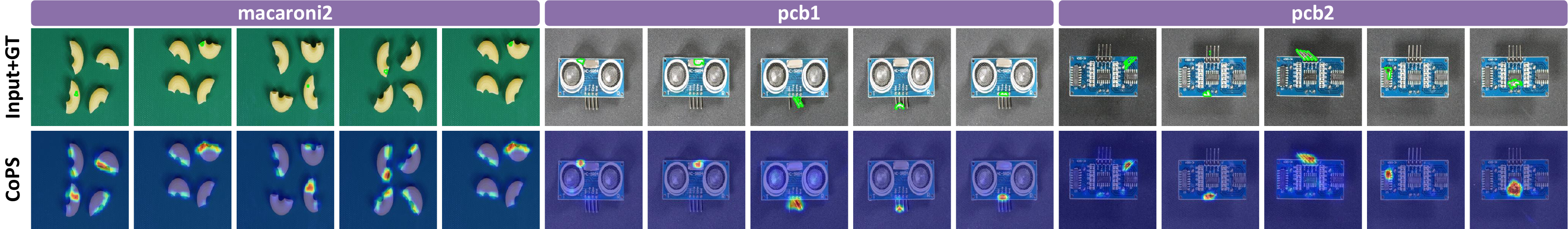}
    \caption{Qualitative segmentation results for the macaroni2, pcb1, and pcb2 categories from the VisA dataset.}
    \label{fig:visa3_}
\end{figure*}

\begin{figure*}[t]
    \centering
    \includegraphics[width=\linewidth]{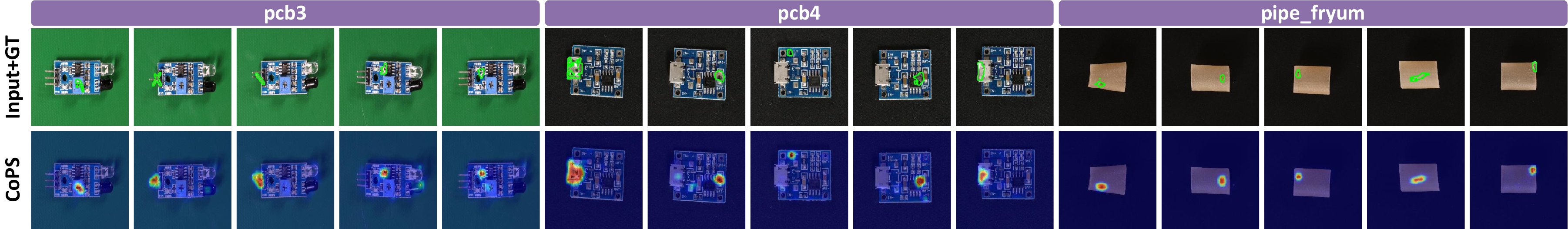}
    \caption{Qualitative segmentation results for the pcb3, pcb4, and pipe fryum categories from the VisA dataset.}
    \label{fig:visa4_}
\end{figure*}

\begin{figure*}[t]
    \centering
    \includegraphics[width=\linewidth]{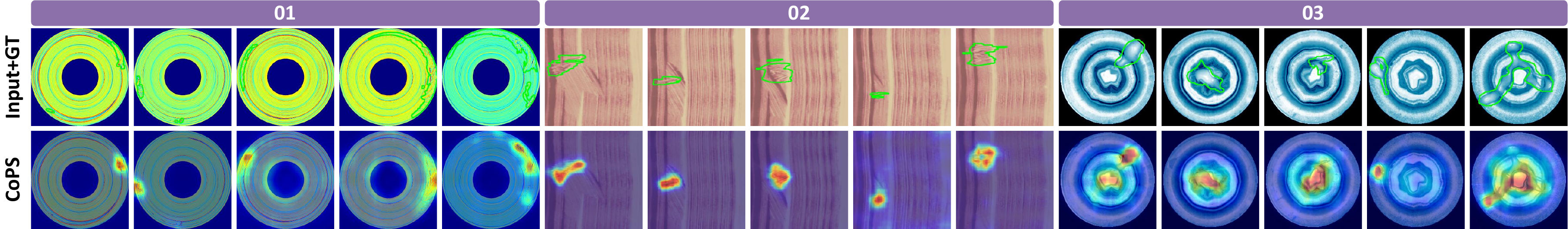}
    \caption{Qualitative segmentation results for the 01, 02, and 03 categories from the BTAD dataset.}
    \label{fig:btad_}
\end{figure*}

\begin{figure*}[t]
    \centering
    \includegraphics[width=\linewidth]{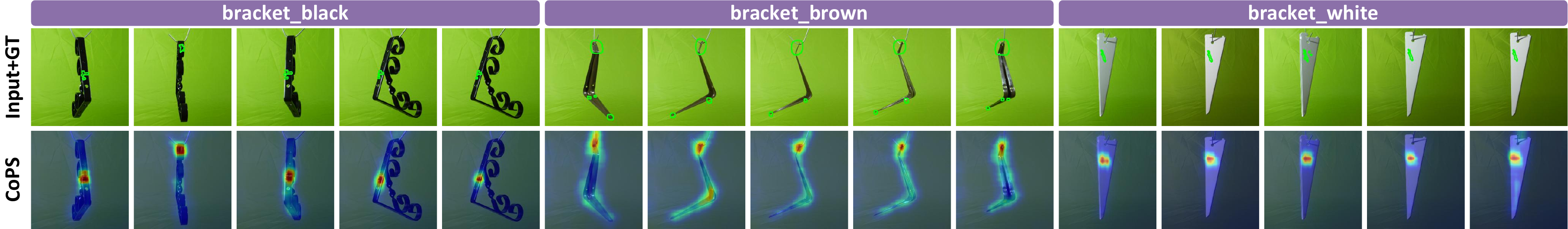}
    \caption{Qualitative segmentation results for the bracket black, brown, and white categories from the MPDD dataset.}
    \label{fig:mpdd1_}
\end{figure*}

\begin{figure*}[t]
    \centering
    \includegraphics[width=\linewidth]{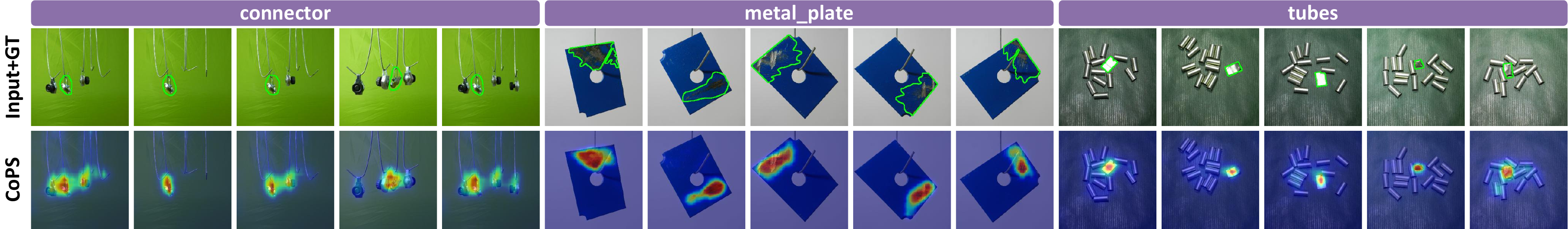}
    \caption{Qualitative segmentation results for the connector, metal plate, and tubes categories from the MPDD dataset.}
    \label{fig:mpdd2_}
\end{figure*}

\begin{figure*}[t]
    \centering
    \includegraphics[width=\linewidth]{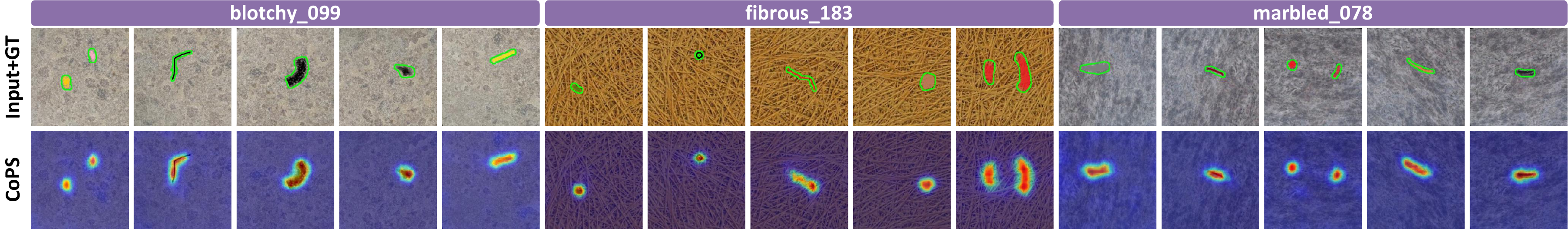}
    \caption{Qualitative segmentation results for the blotchy, fibrous, and marbled categories from the DTD-Synthetic dataset.}
    \label{fig:dtd1_}
\end{figure*}

\begin{figure*}[t]
    \centering
    \includegraphics[width=\linewidth]{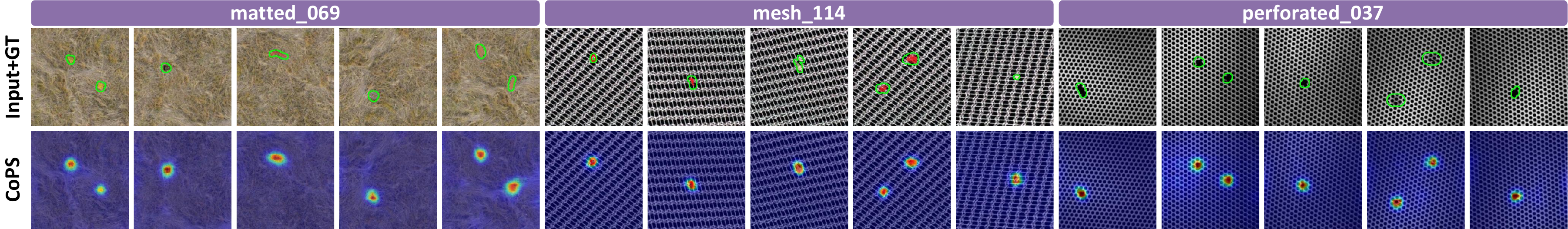}
    \caption{Qualitative segmentation results for the matted, mesh, and perforated categories from the DTD-Synthetic dataset.}
    \label{fig:dtd2_}
\end{figure*}

\begin{figure*}[t]
    \centering
    \includegraphics[width=\linewidth]{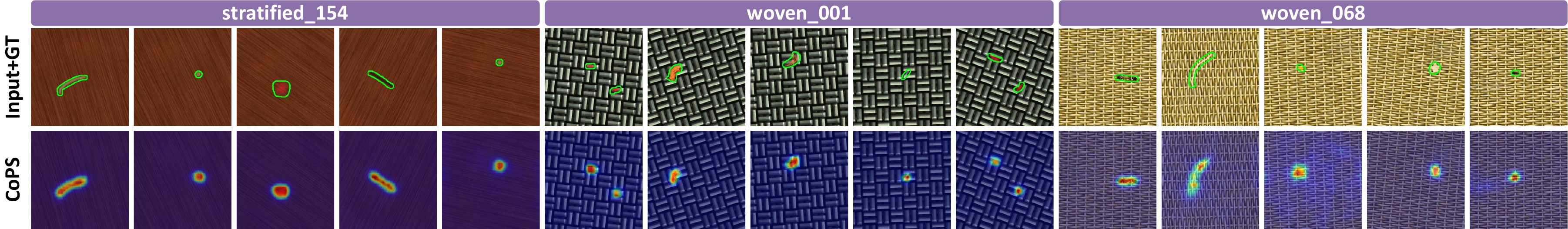}
    \caption{Qualitative segmentation results for the stratified, woven1, and woven2 categories from the DTD-Synthetic dataset.}
    \label{fig:dtd3_}
\end{figure*}

\begin{figure*}[t]
    \centering
    \includegraphics[width=\linewidth]{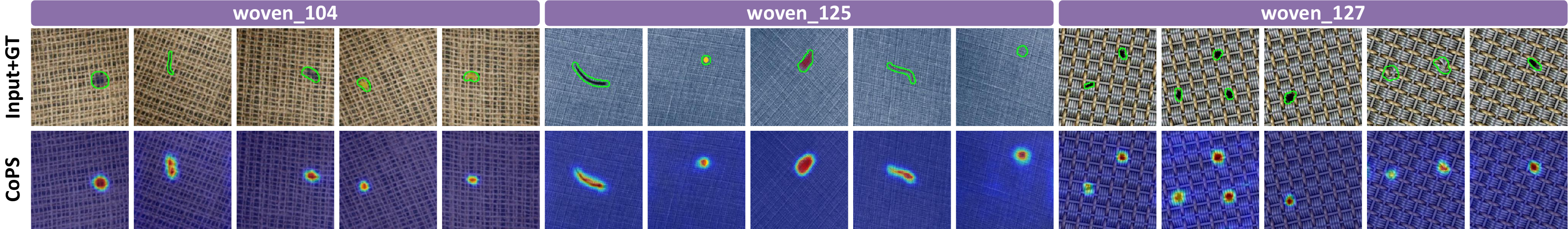}
    \caption{Qualitative segmentation results for the woven3, woven4, and woven5 categories from the DTD-Synthetic dataset.}
    \label{fig:dtd4_}
\end{figure*}

\begin{figure*}[t]
    \centering
    \includegraphics[width=\linewidth]{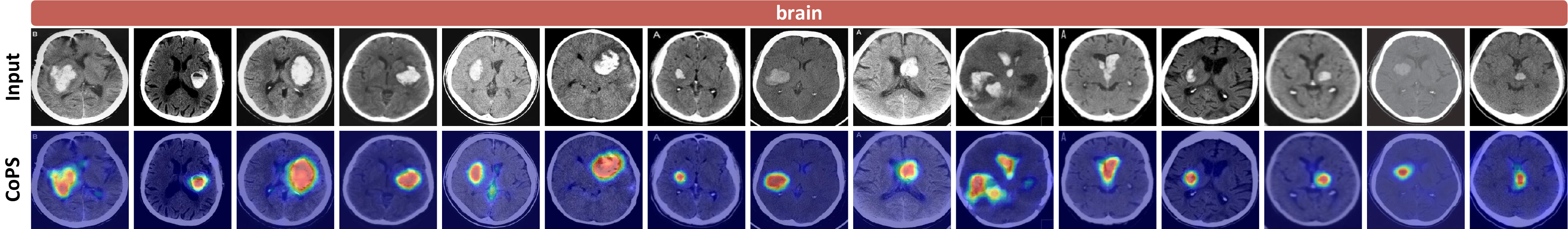}
    \caption{Qualitative segmentation results for the brain category from the HeadCT dataset.}
    \label{fig:headct_}
\end{figure*}

\begin{figure*}[t]
    \centering
    \includegraphics[width=\linewidth]{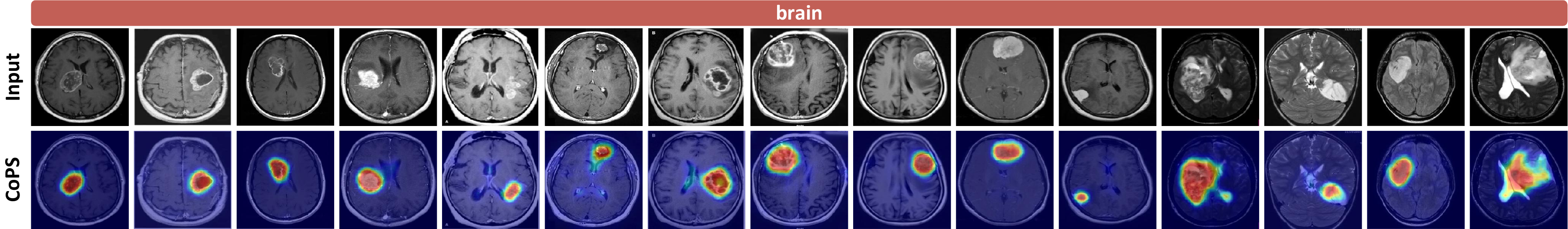}
    \caption{Qualitative segmentation results for the brain category from the BrainMRI dataset.}
    \label{fig:brainmri_}
\end{figure*}

\begin{figure*}[t]
    \centering
    \includegraphics[width=\linewidth]{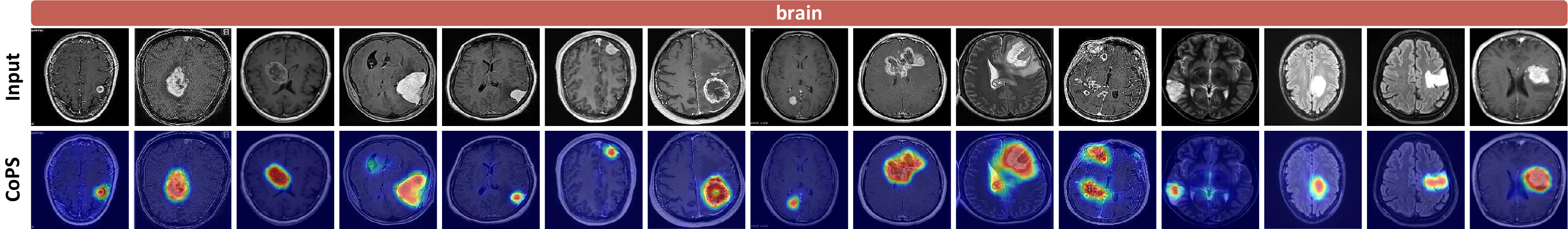}
    \caption{Qualitative segmentation results for the brain category from the Br35H dataset.}
    \label{fig:br35h_}
\end{figure*}

\begin{figure*}[t]
    \centering
    \includegraphics[width=\linewidth]{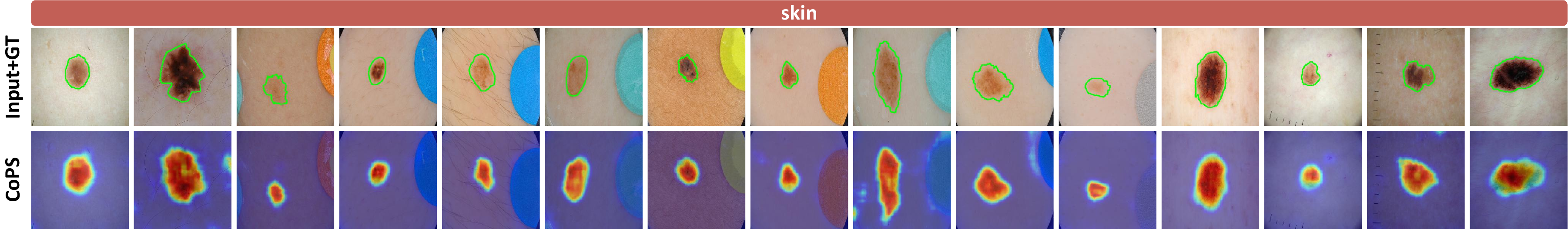}
    \caption{Qualitative segmentation results for the skin category from the ISIC dataset.}
    \label{fig:isic_}
\end{figure*}

\begin{figure*}[t]
    \centering
    \includegraphics[width=\linewidth]{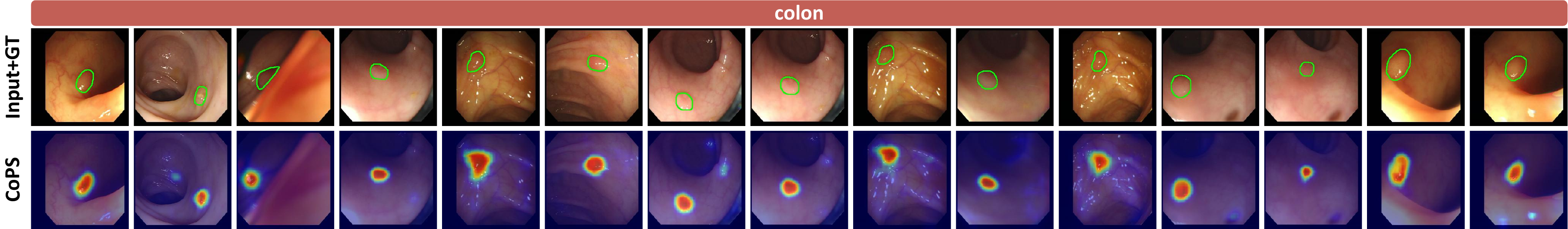}
    \caption{Qualitative segmentation results for the colon category from the CVC-ColonDB dataset.}
    \label{fig:colondb_}
\end{figure*}

\begin{figure*}[t]
    \centering
    \includegraphics[width=\linewidth]{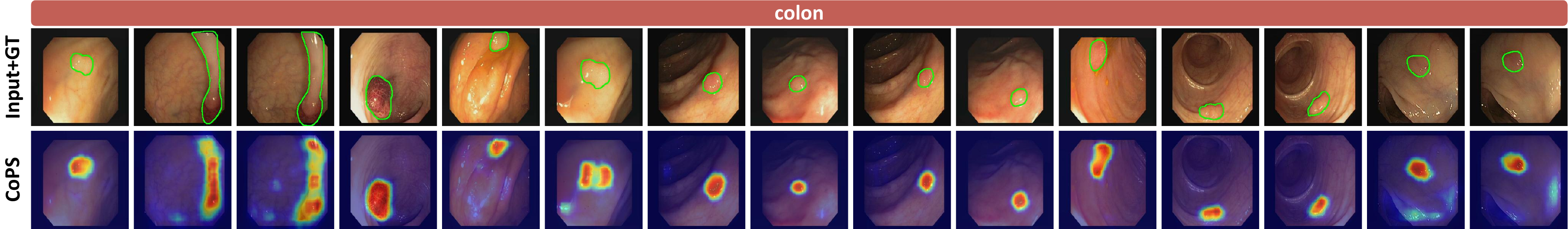}
    \caption{Qualitative segmentation results for the colon category from the CVC-ClinicDB dataset.}
    \label{fig:clinicdb_}
\end{figure*}

\begin{figure*}[t]
    \centering
    \includegraphics[width=\linewidth]{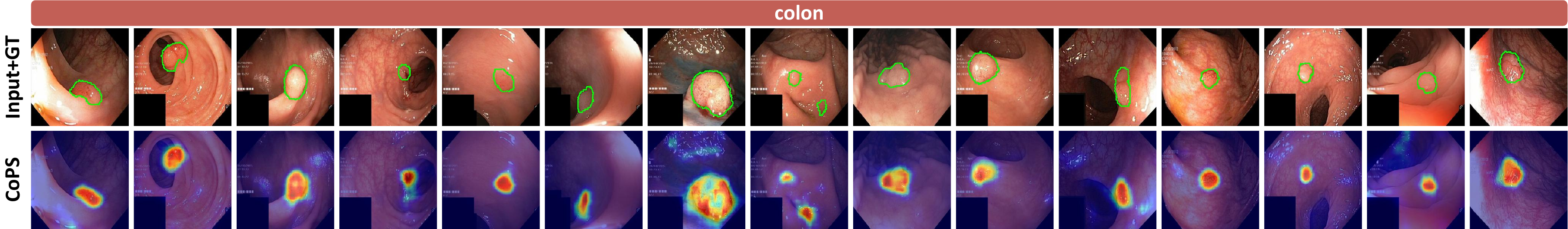}
    \caption{Qualitative segmentation results for the colon category from the Kvasir dataset.}
    \label{fig:kvasir_}
\end{figure*}

\begin{figure*}[t]
    \centering
    \includegraphics[width=\linewidth]{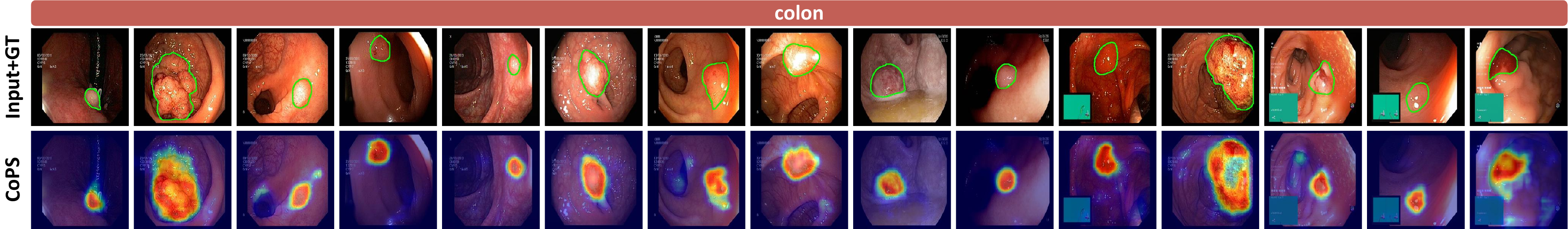}
    \caption{Qualitative segmentation results for the colon category from the Endo dataset.}
    \label{fig:endo_}
\end{figure*}

In this section, we present the visualization results of CoPS across all categories on 13 industrial and medical datasets.
Specifically, Figs~\ref{fig:mvtec1_}--\ref{fig:mvtec5_} show the visualization results of CoPS on all 15 categories of the MVTec-AD dataset.
Figs~\ref{fig:visa1_}--\ref{fig:visa4_} present results on all 12 categories of the VisA dataset.
Fig.~\ref{fig:btad_} displays the visualizations for all 3 categories in BTAD.
Figs~\ref{fig:mpdd1_}--\ref{fig:mpdd2_} illustrate results on all 6 categories of MPDD.
Figs~\ref{fig:dtd1_}--\ref{fig:dtd4_} cover all 12 categories of DTD-Synthetic.
Figs.~\ref{fig:headct_}--\ref{fig:endo_} show comprehensive visualizations for all categories
included in the HeadCT, BrainMRI, Br35H, ISIC, CVC-ColonDB, CVC-ClinicDB, Kvasir, and Endo datasets.
It is important to note that HeadCT, BrainMRI, and Br35H (Figs.~\ref{fig:headct_}--\ref{fig:br35h_}) do not provide pixel-level annotations.
As a result, only the input images can be displayed, and no ground-truth contours are available for qualitative comparison.

\clearpage

\end{document}